%% file: root.tex
\documentclass[letterpaper, 10 pt, conference]{ieeeconf}  
\usepackage{cite}

\IEEEoverridecommandlockouts                              

\usepackage{graphicx} 
\usepackage{amsmath} 
\usepackage{float}
\usepackage{hyperref}
\hypersetup{colorlinks,allcolors=black}
\usepackage{color,soul}
\usepackage{xcolor}

\newcommand\highlightReference[1]{%
  \expandafter\newcommand\csname highlightReference-#1\endcsname{}%
}
\let\oldbibitem\bibitem
\def\bibitem#1 #2\par{%
  \expandafter\ifx\csname highlightReference-#1\endcsname\relax
    \oldbibitem{#1}#2\par
  \else
    \oldbibitem{#1}\highlight{#2}\par
  \fi
}
\usepackage{color,soul}
\newcommand\highlight[1]{\hl{#1}}

\title{\LARGE \bf
Integration of a Variable Stiffness Link for Long-Reach Aerial Manipulation}

 \author{Manuel J. Fernandez$^{1}$, Alejandro Suarez$^{2}$, Anibal Ollero$^{2}$ and Matteo Fumagalli$^{1}$  
 \thanks{$^1$ Department of Electrical and Photonics Engineering Control, Robotics and Embodied AI, Technical University of Denmark, 2800 Kongens Lyngby, Denmark, {\tt\small mjfgo@dtu.dk}}%
 \thanks{$^2$ GRVC Robotics Laboratory, Universidad de Sevilla, 41092 Seville, Spain}%
 \thanks{Funded by the European Union. Views and opinions expressed are however those of
the author(s) only and do not necessarily reflect those of the European Union or
European Commission. Neither the European Union nor the granting authority can be
held responsible for them}%
 }

\begin{document}

\maketitle
\thispagestyle{empty}
\pagestyle{empty}

\begin{abstract}
This paper presents the integration of a Variable Stiffness Link (VSL) for long-reach aerial manipulation, enabling adaptable mechanical coupling between an aerial multirotor platform and a dual-arm manipulator. Conventional long-reach manipulation systems rely on rigid or cable connections, which limit precision or transmit disturbances to the aerial vehicle. The proposed VSL introduces an adjustable stiffness mechanism that allows the link to behave either as a flexible rope or as a rigid rod, depending on task requirements.

The system is mounted on a quadrotor equipped with the LiCAS dual-arm manipulator and evaluated through teleoperated experiments, involving external disturbances and parcel transportation tasks. Results demonstrate that varying the link stiffness significantly modifies the dynamic interaction between the UAV and the payload. The flexible configuration attenuates external impacts and aerodynamic perturbations, while the rigid configuration improves positional accuracy during manipulation phases.

These results confirm that VSL enhances versatility and safety, providing a controllable trade-off between compliance and precision. Future work will focus on autonomous stiffness regulation, multi-rope configurations, cooperative aerial manipulation and user studies to further assess its impact on teleoperated and semi-autonomous aerial tasks.
\end{abstract}

\input{sections/a_intro}


\input{sections/m_system}

\input{sections/n_model}

\input{sections/x_experiments}


\input{sections/z_conclusion}

\input{sections/zz_acknowledgement}

\addtolength{\textheight}{-12cm}   

\highlightReference{vsl_depei}
\bibliographystyle{IEEEtran}
\bibliography{bibliography}

\end{document}

%% file: sections/a_intro.tex
\section{INTRODUCTION}\label{sec:intro}
Aerial robotic manipulation extends the capabilities of Unmanned Aerial Vehicles (UAVs) from passive observation to active physical interaction with the environment. Applications such as inspection \cite{jordan2018state}, maintenance \cite{article}, and transportation \cite{11007930} require UAVs not only to perceive but also to exert forces safely and accurately. Over the past decade, aerial manipulators have evolved from basic gripper-equipped multirotors \cite{heredia2014control} to sophisticated systems capable of dynamic physical interaction \cite{li2025six}. A comprehensive overview of this evolution is provided in \cite{past_present}.

Among different designs, long-reach aerial manipulators \cite{9088973} extend the operational workspace of UAVs, allowing interaction with distant or hazardous targets while maintaining a safe separation between the aerial platform and the environment. Originally developed for inspection and maintenance of high-voltage power infrastructures \cite{9469805, 9836039, nekoo2023constrained}, this design also benefits applications such as parcel delivery \cite{8594123} and remote human–robot interaction \cite{9836039}. However, existing long-reach systems rely mainly on suspended cables \cite{9088973}, presenting intrinsic limitations.

Cable-suspended manipulators effectively isolate the aerodynamic disturbances generated by the rotors and reduce collision risk, but they typically suffer from pendulum-like oscillations (\cite{8743463}, \cite{10611383}) that complicate precise control. Conversely, the use of a long rigid extension could improve precision but directly transmit contact forces and external disturbances to the UAV, compromising stability. This fundamental trade-off between compliance and precision remains an open challenge for long-reach aerial manipulation. None of the existing long-reach aerial manipulators provides adjustable stiffness to actively modulate the dynamic coupling between the UAV and the payload.

To address this challenge, this paper proposes integrating a Variable Stiffness Link (VSL)\cite{vsl_depei} between the UAV and the manipulator. The VSL dynamically adjusts its stiffness during operation, i.e. the mechanical coupling, allowing the connection to behave as either a compliant rope or a rigid rod, depending on task requirements. This adaptability allows compliant behavior during transportation phases and improved accuracy during grasping or contact interactions. Although variable stiffness mechanisms have been explored in robotics \cite{stiffness, VSL}, to the best of the authors’ knowledge, this is the first integration of such a device into an aerial manipulation system.

\begin{figure}[!ht]
    \centering
    \includegraphics[width=1.0\linewidth]{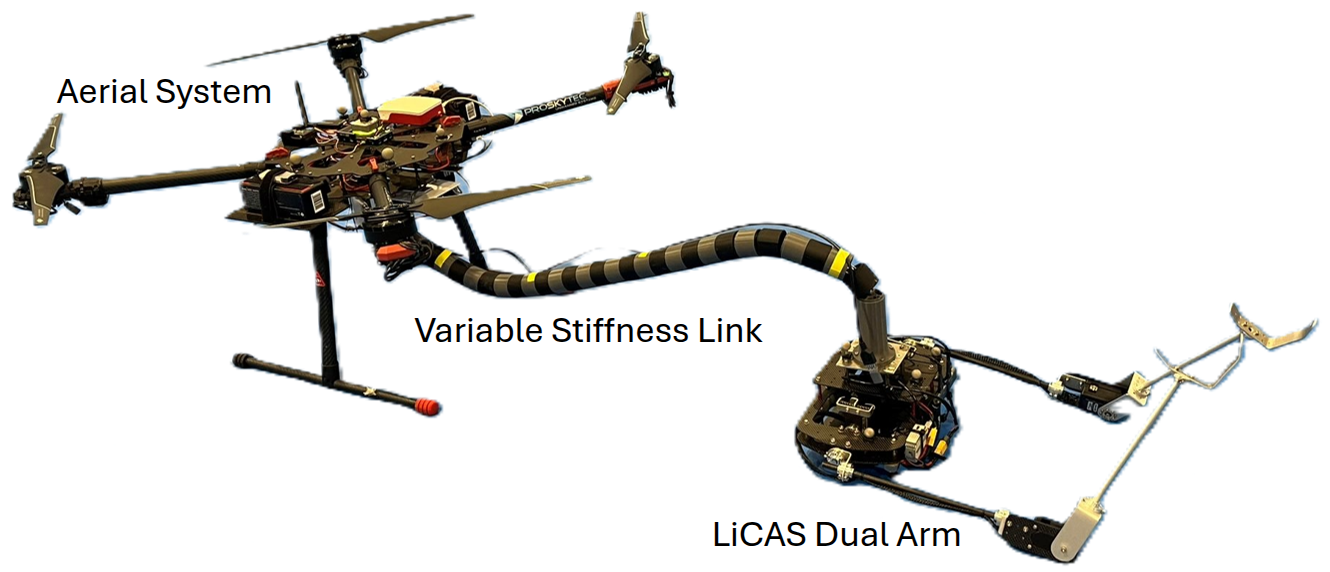}
    \caption{Aerial manipulator - VSL connects the aerial system with the LiCAS dual-arm manipulator}
    \label{fig:aerial_manipulator}
\end{figure}

The main contributions of this work are:
\begin{itemize}
    \item The integration of a lightweight, cable-driven Variable Stiffness Link for aerial manipulation.
    
    \item The experimental validation of the VSL under different stiffness configurations, including external disturbance tests and parcel transportation.
    
    \item Demonstration of teleoperated manipulation with the VSL, with and without the dual-arm LiCAS system attached, showing the feasibility of long-reach operations with dynamic stiffness adjustment.
\end{itemize}

The rest of this paper is structured as follows: Section \ref{sec:sys_over} presents the system overview, followed by the dynamic modeling in Section \ref{sec:model}. The experimental results are shown in Section \ref{sec:experiments}, and Section \ref{sec:conclusions} concludes the paper and outlines future research directions.

%% file: sections/m_system.tex
\section{SYSTEM OVERVIEW}\label{sec:sys_over}
\subsection{Aerial system}
The aerial platform is a custom quadrotor based on the Tarot X4 (Fig \ref{fig:aerial_manipulator}), equipped with a Pixhawk flight controller running Ardupilot firmware and a Raspberry Pi computer with Ubuntu 20.04 OS. The total weight of the vehicle, including two 6S batteries, is 5.4kg. The flight controller is manually piloted in position-hold mode during all experiments, while the onboard computer handles high-level commands and data acquisition.


\subsection{Variable Stiffness Link}
\begin{figure}[!ht]
    \centering
    \includegraphics[width=1.0\linewidth]{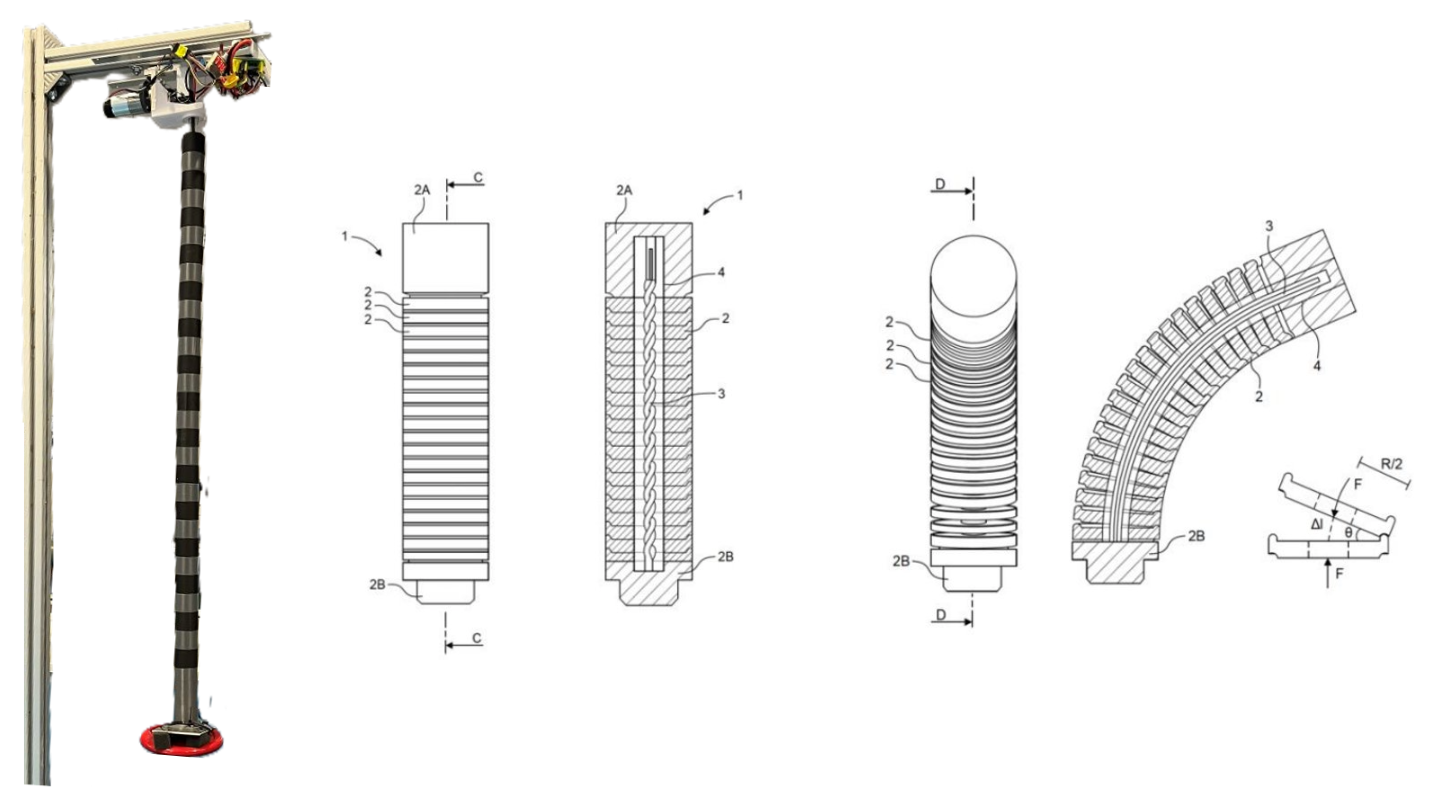}
    \caption{VSL mounted on a workbench with a testing weight on the tip (left), sketch of VSL in rest configuration (middle) and bent state (right)}
    \label{fig:vsl_description}
\end{figure}

The Variable Stiffness Link (VSL) is a 1m-long custom-built element that connects the UAV with the manipulator.The link consists of rigid, vertebra-like segments fabricated in PETG and connected in series through internal cables. The total weight of the assembly (Fig. \ref{fig:vsl_description}), including the electronics and mounting plate, is 1.2 kg.

Stiffness is adjusted through a cable-driven mechanism. A GB37Y3530-12V-90E DC geared motor, controlled by an Arduino Mega 2560 PRO board and a Pololu md31c driver, twists the internal ropes to vary the link’s rigidity, as shown in Fig \ref{fig:vsl_description}. The Arduino executes a PID controller that regulates the motor position based on encoder feedback, achieving transitions between flexible and rigid configurations.

This setup includes a load cell SEN-HX711-20 installed between the drone frame and the VSL to estimate the pulling strength that is being carried out.

\begin{figure}[!ht]
    \centering
    \includegraphics[width=1.0\linewidth]{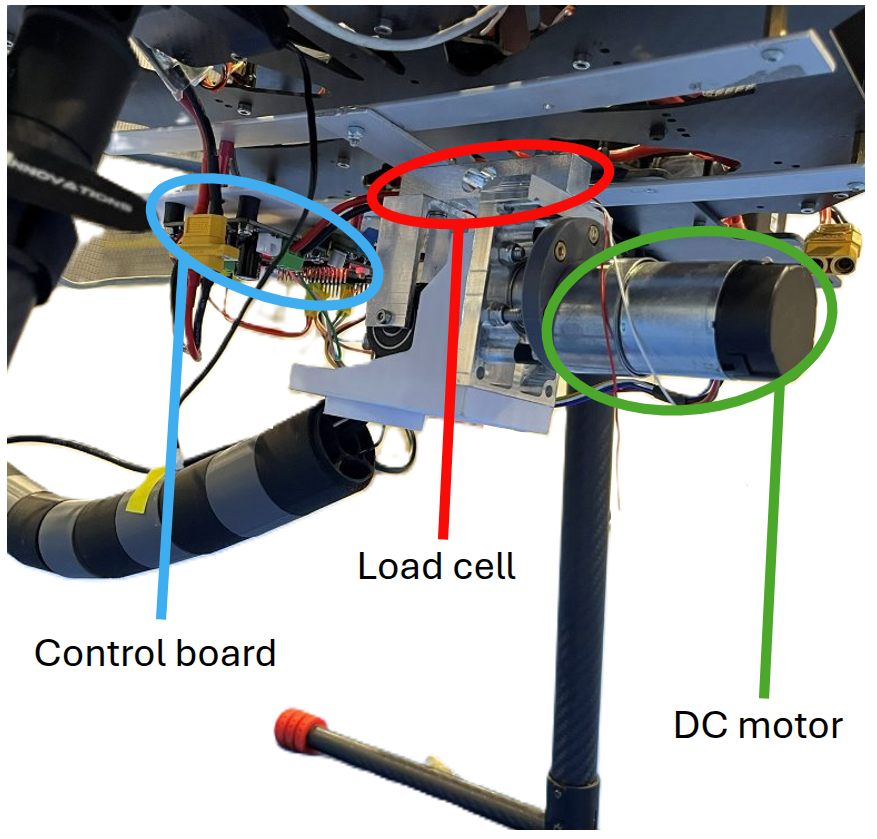}
    \caption{Mounting point detail of VSL under the aerial system}
    \label{fig:vsl_attachment}
\end{figure}

A load cell (SEN-HX711-20) is mounted between the UAV frame and the link (Fig. \ref{fig:vsl_attachment}) to measure the tension transmitted through the VSL. This sensor is used to estimate the pulling effort and to study force transmission during manipulation tasks.

\subsection{Manipulator}
The end-effector is the Lightweight and Compliant Anthropomorphic Dual-Arm System (LiCAS) A1, shown in Fig. \ref{fig:aerial_manipulator}. It is attached to the tip of the VSL and connected to the onboard computer. Each arm has two Degrees of Freedom and human-like proportions, with a maximum reach of 0.5 m and an inter-arm distance of 0.36 m. The manipulator weighs approximately 2 kg, including its batteries and a DJI O3 FPV camera for visual feedback.

The system uses a leader–follower teleoperation architecture, where an external LiCAS AC1 acts as the master device manipulated by a human operator (Fig. 5). The master’s joint positions are sent via UDP sockets to the onboard computer, which drives the follower manipulator. The operator receives real-time video feedback through DJI Goggles 2 connected to the FPV camera.

\begin{figure}[!ht]
    \centering
    \includegraphics[width=1.0\linewidth]{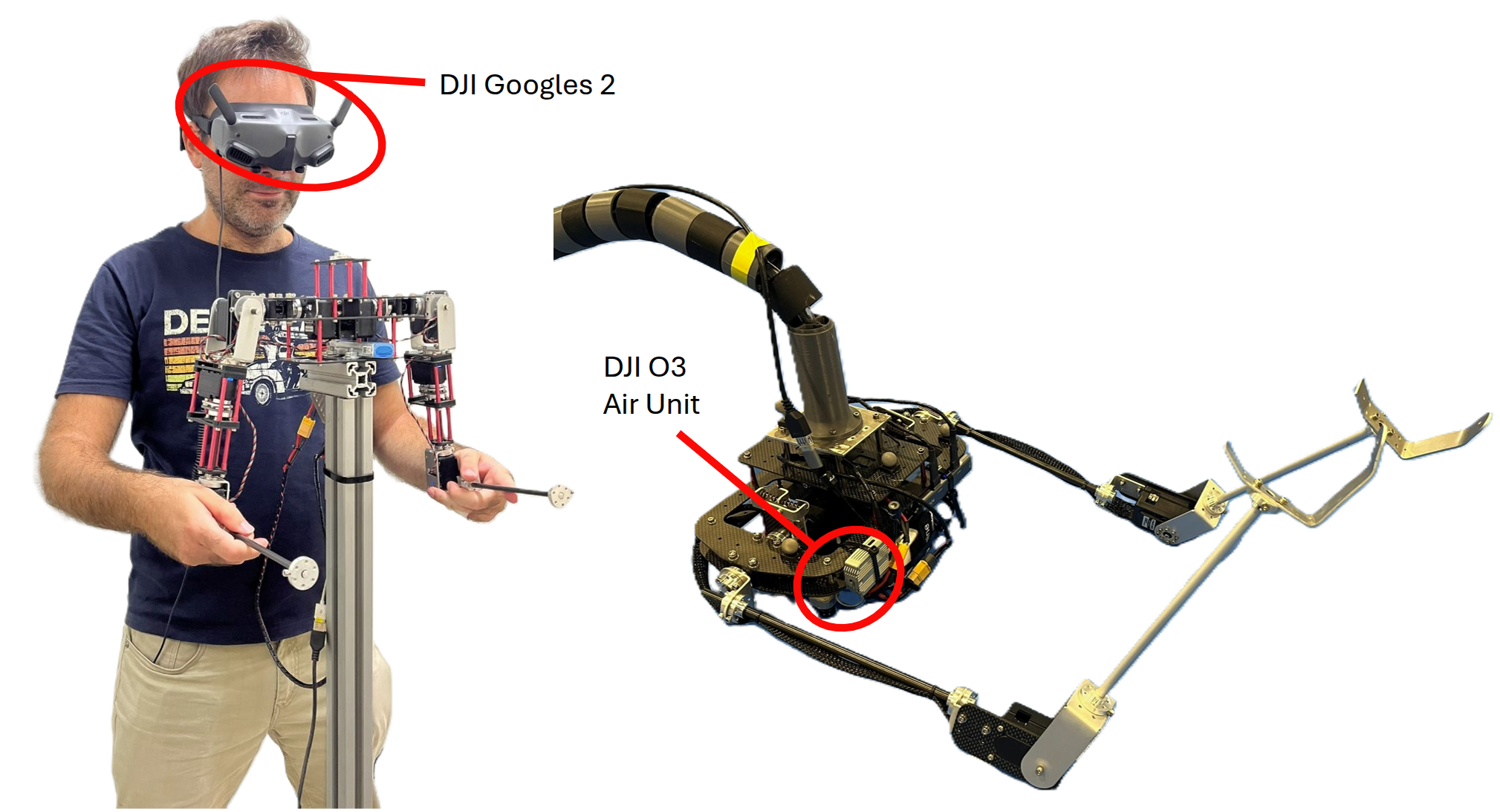}
    \caption{Operator with LiCAS AC1 teleoperation system (left) and dual-arm manipulator LiCAS A1 mounted on the aerial system (right)}
    \label{fig:teleoperation_device}
\end{figure}

\subsection{System architecture}
\begin{figure}[!ht]
    \centering
    \includegraphics[width=1.0\linewidth]{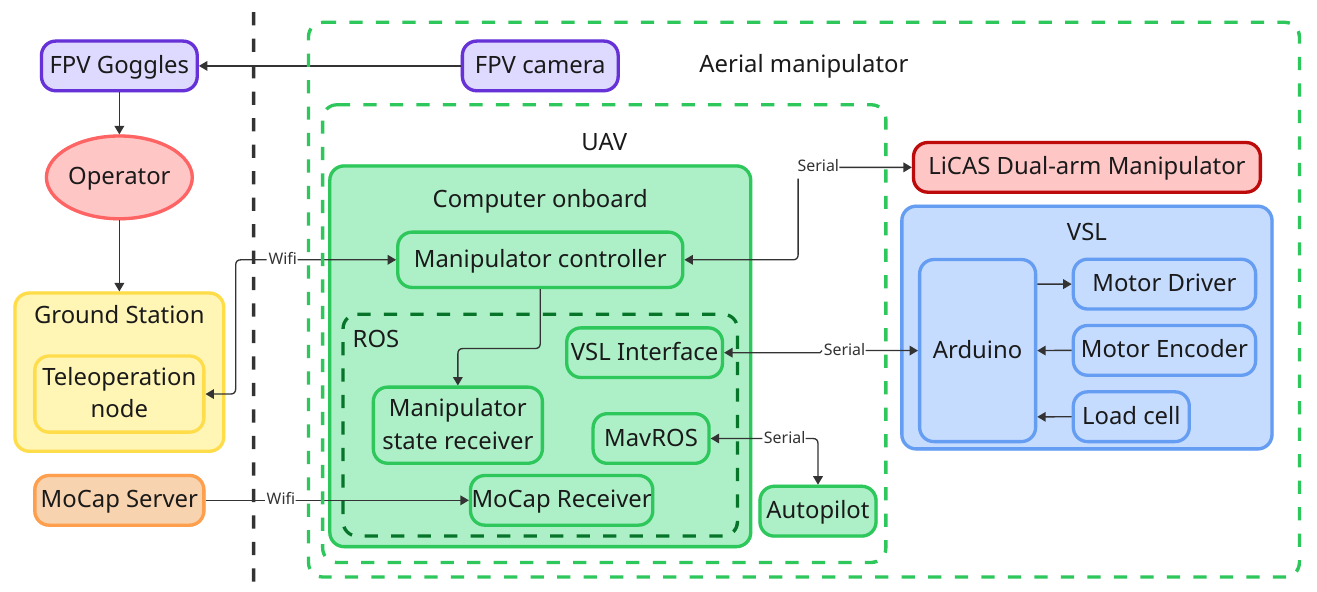}
    \caption{System architecture diagram}
    \label{fig:diagram}
\end{figure}

Fig. \ref{fig:diagram} illustrates the overall architecture and communication flow. The onboard computer runs ROS Noetic, managing all nodes and ensuring timestamp synchronization. It exchanges data with the Pixhawk autopilot through MAVROS to close the UAV position loop.

An OptiTrack motion-capture system provides external position feedback for both the UAV and the VSL tip. The VSL interface node allows stiffness setpoint commands to the Arduino controller and publishes motor encoder and load-cell data feedback. The manipulator controller receives reference commands from the teleoperation node and returns state information to the ground station through UDP communication.

The FPV camera streams directly to the operator’s goggles, providing a first-person perspective during experiments.

\subsection{Control Strategy}
During all experiments, the UAV operates in position-hold mode, controlled by the Ardupilot flight stack running on the Pixhawk. The onboard computer communicates with the flight controller through MAVROS, closing the position loop using the external motion-capture feedback. The safety pilot supervises the flight and manually commands horizontal displacement when required for the task.

The Variable Stiffness Link (VSL) is adjusted by sending position setpoints to the Arduino controller from a ROS node. Each command defines a motor position corresponding to a target stiffness level:

\begin{itemize}
    \item Flexible mode corresponds to minimal rope twisting, allowing compliant motion,
    \item Rigid mode corresponds to the maximum safe twist angle before material saturation.
\end{itemize}

A PID controller on the Arduino regulates the motor position to track these setpoints, ensuring smooth transitions between configurations. The link state (target and measured stiffness, encoder position, and load-cell data) is continuously published by VSL Interface as a ROS topics, allowing to be synchronized with UAV and manipulator measurements.

The LiCAS manipulator is teleoperated in a leader–follower scheme. The master manipulator reproduces human motion, and the corresponding joint commands are streamed to the onboard follower through UDP. The UAV maintains position autonomously while the operator controls the manipulator, allowing direct evaluation of how different VSL stiffness levels affect dynamic coupling and manipulation accuracy.

%% file: sections/n_model.tex
\section{MODELING}\label{sec:model}
The system consist of a quadrotor (mass $m_q$, inertia $J=\mathrm{diag}(J_x,J_y,J_z)$) carrying a payload mass \(m_p\) at the end of the Variable Stiffness Link (VSL) of length $l$, attached at the Center of Mass (CoM) of the aerial vehicle. The actuator executes a torsional force pulling from the tip of the VSL to modulate the rigidity of the link. The VSL behaves as a flexible cable or a rigid rod, modifying the dynamic coupling between the payload and the UAV.

Considering the UAV in a hovering state the autopilot attitude controller will control the platform keeping a fixed zero target setpoint for pitch and roll $\theta:=[\phi\ \theta]^T$). This state can be modeled as a virtual virtual spring-damper system:

\begin{equation}
J_{xy}\ddot{\theta}+D_c\dot{\theta}+K_c\theta=\tau_{\mathrm{ext}},
\label{eq:att-loop}
\end{equation}

where $K_c\simeq K_pJ_{xy}$ and $D_c\simeq K_dJ_{xy}$ are the virtual stiffness/damping and $\tau_{\mathrm{ext}}$ are external torques, including the VSL reaction and other disturbances. This statement assumes the flexible link transmits force but no moment about the UAV roll/pitch axes. The VSL is coupled from the UAV in this state.

Considering $\alpha:=[\alpha_x\ \alpha_y]^T$ deviations of the VSL tip in the roll and pitch respect to the UAV, a pendular motion appears while the VSL is in motion. This behave, per axis, correspond to:
\begin{equation}
m_pl^2\ddot{\alpha}+c_p\dot{\alpha}+m_pgl\,\alpha
+ k_s(\alpha-\theta) + c_s(\dot{\alpha}-\dot{\theta}) = \tau_d,
\label{eq:pend}
\end{equation}
where $c_p$ is the damping from the pendulum, and $\tau_d$ is an external torque (e.g., impact or wind) at the tip. $k_s$ and $c_s$ are the effective stiffness and dampingat the UAV/VSL attachment, i.e. zero when VSL is flexible. 

The variation of stiffness is considered with $\sigma\in[0,1]$ (0 = fully flexible, 1 = fully rigid):
\[
k_s(\sigma)=\sigma k_{\max},\qquad c_s(\sigma)=\sigma c_{\max}.
\]

During the rigid configuration ($\sigma>0$) the motion of the VSL tip produces a torque on the UAV: 
\begin{equation}
\tau_{\mathrm{vsl}} = -k_s(\theta-\alpha) - c_s(\dot{\theta}-\dot{\alpha}),
\label{eq:tauvsl}
\end{equation}

so that \eqref{eq:att-loop} becomes:
\begin{equation}
J\,\ddot{\theta} + (D_c+c_s)\dot{\theta} + (K_c+k_s)\theta
= k_s\alpha + c_s\dot{\alpha} + \tau_w,
\end{equation}

where $\tau_w$ comes from other external disturbances. 

Thus, the coupled linear time-varying system (via $\sigma$) for the whole system, together with \eqref{eq:pend}, is: 
\begin{align}
J\ddot{\theta} + (D_c+c_s)\dot{\theta} + (K_c+k_s)\theta
&= k_s\alpha + c_s\dot{\alpha} + \tau_w, \\
m_pl^2\ddot{\alpha} + (c_p+c_s)\dot{\alpha} + (m_pgl+k_s)\alpha
&= k_s\theta + c_s\dot{\theta} + \tau_d .
\end{align}

\begin{figure}[!ht]
    \centering    \includegraphics[width=1.0\linewidth]{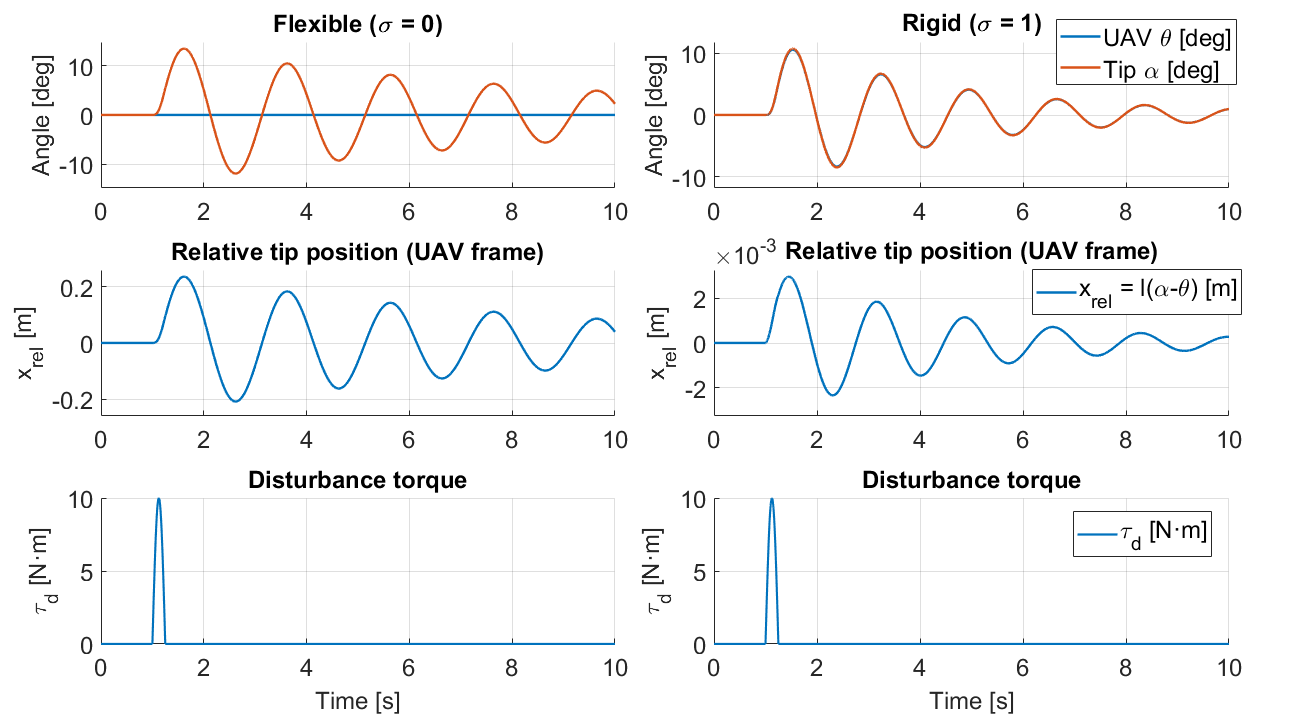}
    \caption{Results simulating the coupled linear time-varying system model under disturbances with rigid (right plots) and flexible (left plots) configuration.}
    \label{fig:model_sim}
\end{figure}

This work considers the study of the two limit configurations: flexible ($\sigma=0$), where attitude and pendulum are dynamics decouple; and rigid ($\sigma=1$) where $\theta$ and $\alpha$ are stiffly coupled and disturbances at the tip are propagated to UAV attitude. Using idealistic parameters, the model is simulated (\ref{fig:model_sim}) to validate the expected behavior of the VSL for the experimental results. The result expose that flexible configuration shows a clear lateral deviation respect to the vehicle while the UAV keeps its attitude, and rigid configuration stays near zero ($\alpha \approx \theta $), while the angles of UAV and VLS  move together due to coupling.
The simulation just consider one axis for simplicity.

%% file: sections/x_experiments.tex
\section{EXPERIMENTS RESULTS}\label{sec:experiments}



Position and attitude for both the UAV and the VSL tip are tracked with an OptiTrack Motion Capture system. The VSL stiffness is controlled by commanding the motor position. Flexible and rigid VSL configuration are highlighted in the plots using blue and red background, respectively. The LiCAS dual-arm manipulator weighs about 2kg and is mounted at the VSL tip in the relevant trials. 

\begin{figure}[!ht]
    \centering    \includegraphics[width=1.0\linewidth]{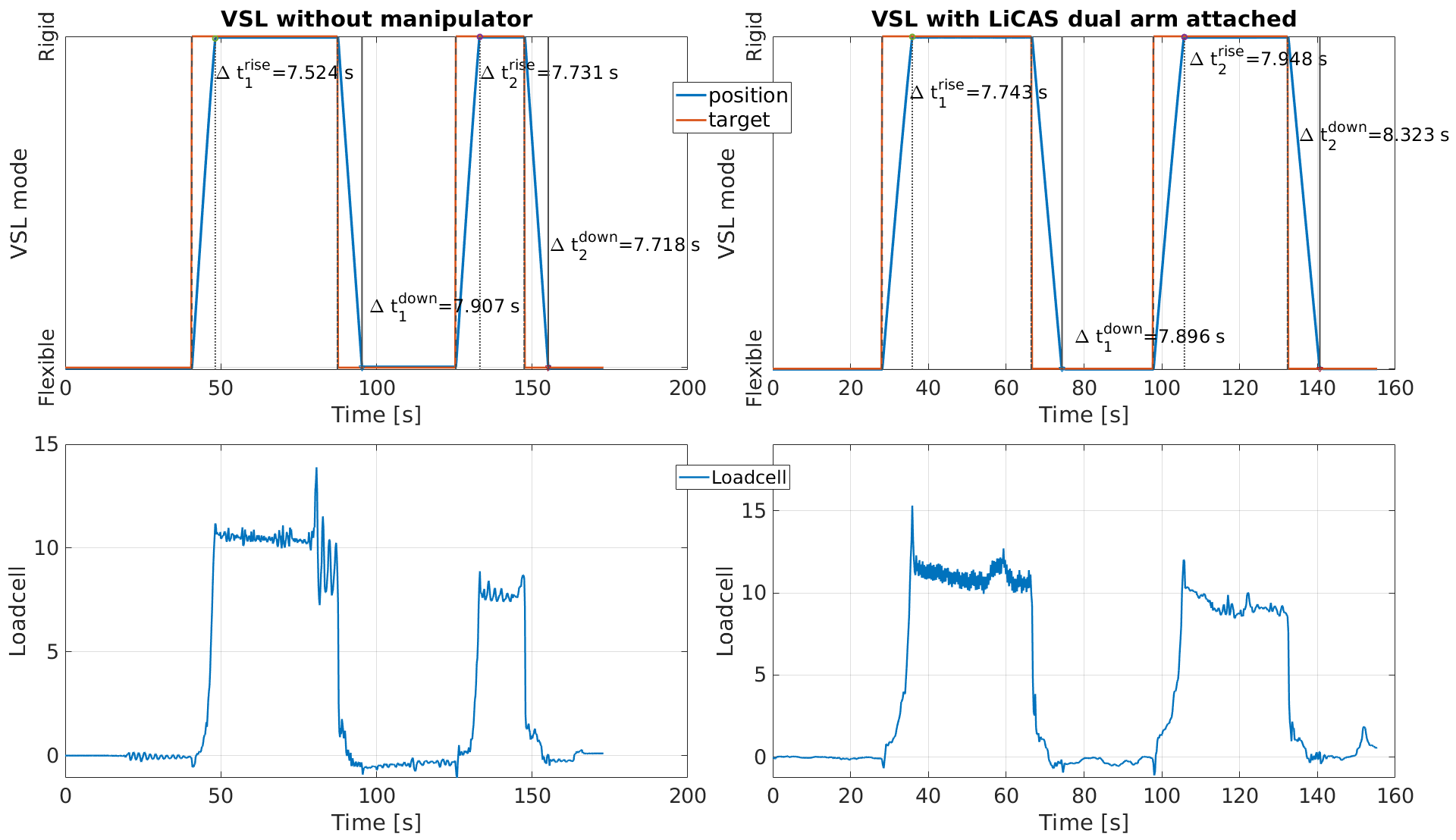}
    \caption{Flexible to rigid transition timing with and without 2kg payload. Target and measured stiffness signals are shown. Transition time about 7.8s. Loadcell measurements registered from their corresponding flight on the top.}
    \label{fig:vsl_state}
\end{figure}

Transition time between stiffness states is approximately 7.8s under all payload conditions. Figure \ref{fig:vsl_state} shows the data from two different flights with and without a 2kg manipulator as payload at the VSL tip. The measured transition time is similar in both cases, which indicates that the actuation can switch configuration reliably under moderate payload. The loadcell measurements mainly reflect the internal effort of the VSL to maintain the rigid configuration rather than the payload lifting force.

\subsection{Disturbance rejection}

The experiments begin by analyzing the dynamic response of the Variable Stiffness Link (VSL) to external disturbances while the UAV hovers in position\footnote{Impact as external disturbance: \href{https://www.youtube.com/watch?v=0EsWP7wj1IM}{youtube.com/watch?v=0EsWP7wj1IM}}. A 2kg weight plate is attached to the tip of the VSL simulating the weight of the manipulator. The VSL is manually configured as flexible or rigid during the experiments.

\begin{figure}[!ht]
    \centering    \includegraphics[width=1.0\linewidth]{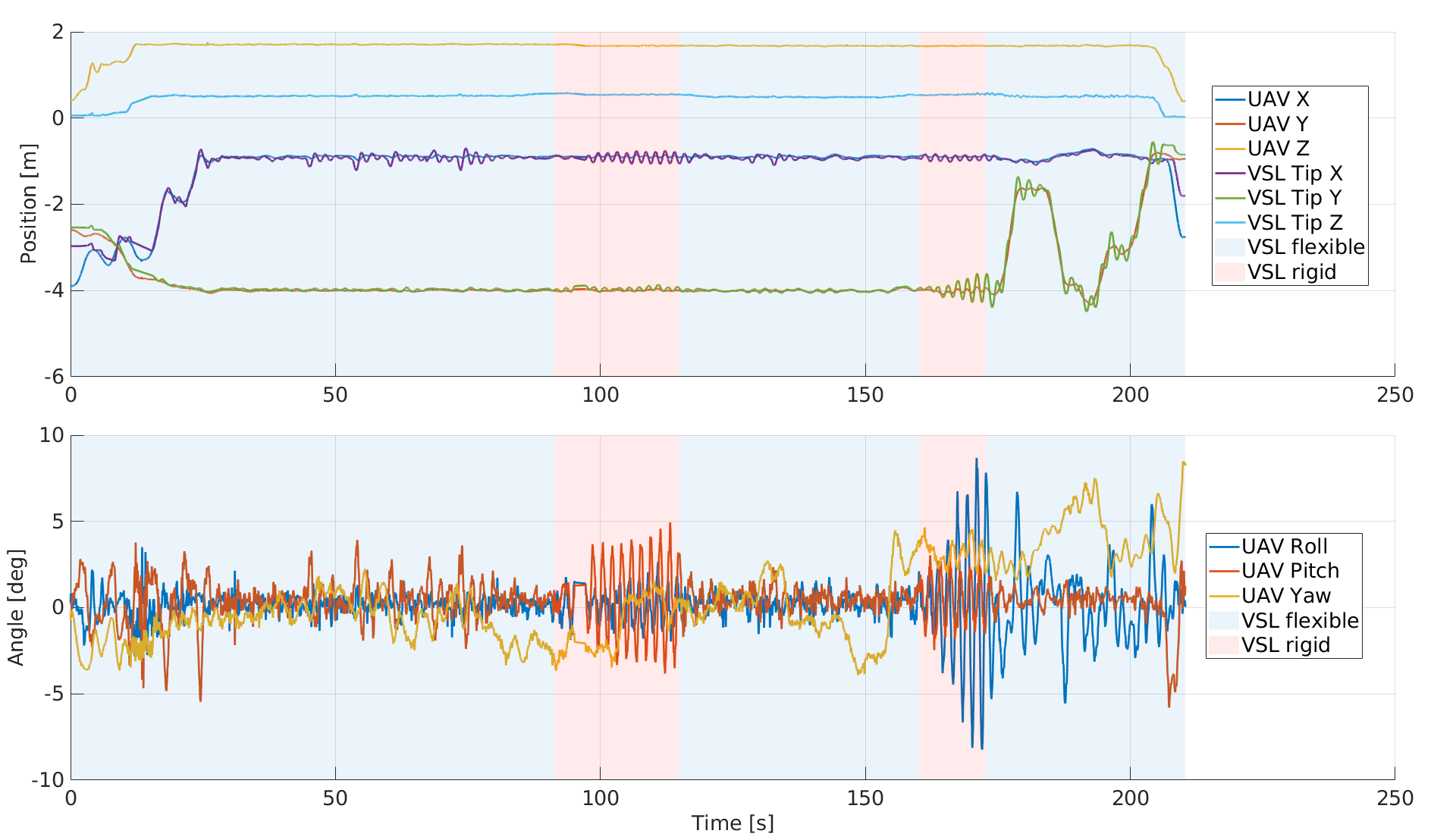}
    \caption{Response to external impacts as disturbances while hovering. Top: UAV and VSL tip positions. Bottom: UAV attitude. Blue is flexible VSL, red is rigid VSL.}
    \label{fig:exp_hits}
\end{figure}

\begin{figure}[!ht]
    \centering    \includegraphics[width=1.0\linewidth]{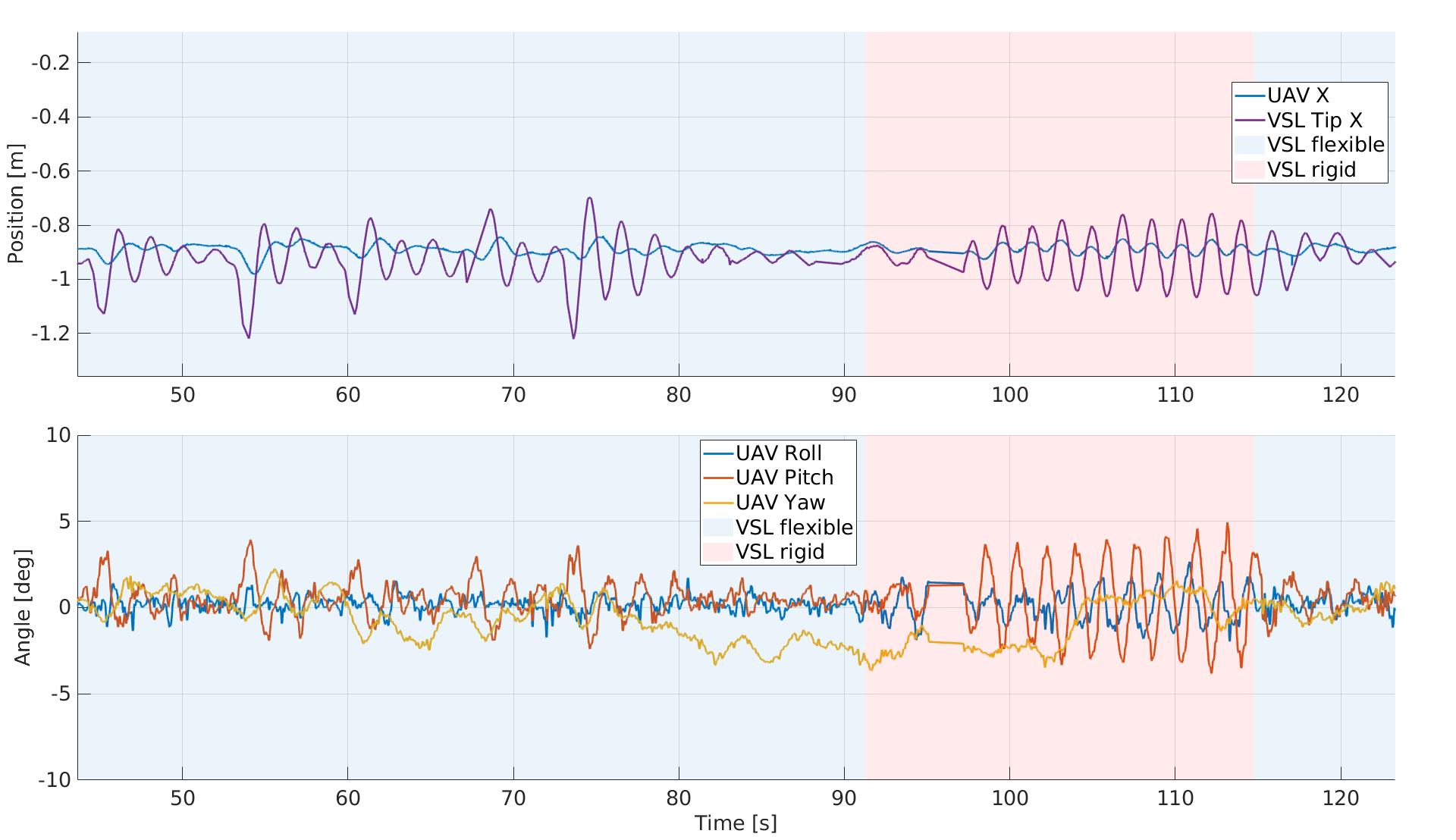}
    \caption{Detailed area from Figure \ref{fig:exp_hits} highlighting a different behave depending on the stiffness.}
    \label{fig:hitting_flex_detail}
\end{figure}

\begin{figure}[!ht]
    \centering    \includegraphics[width=1.0\linewidth]{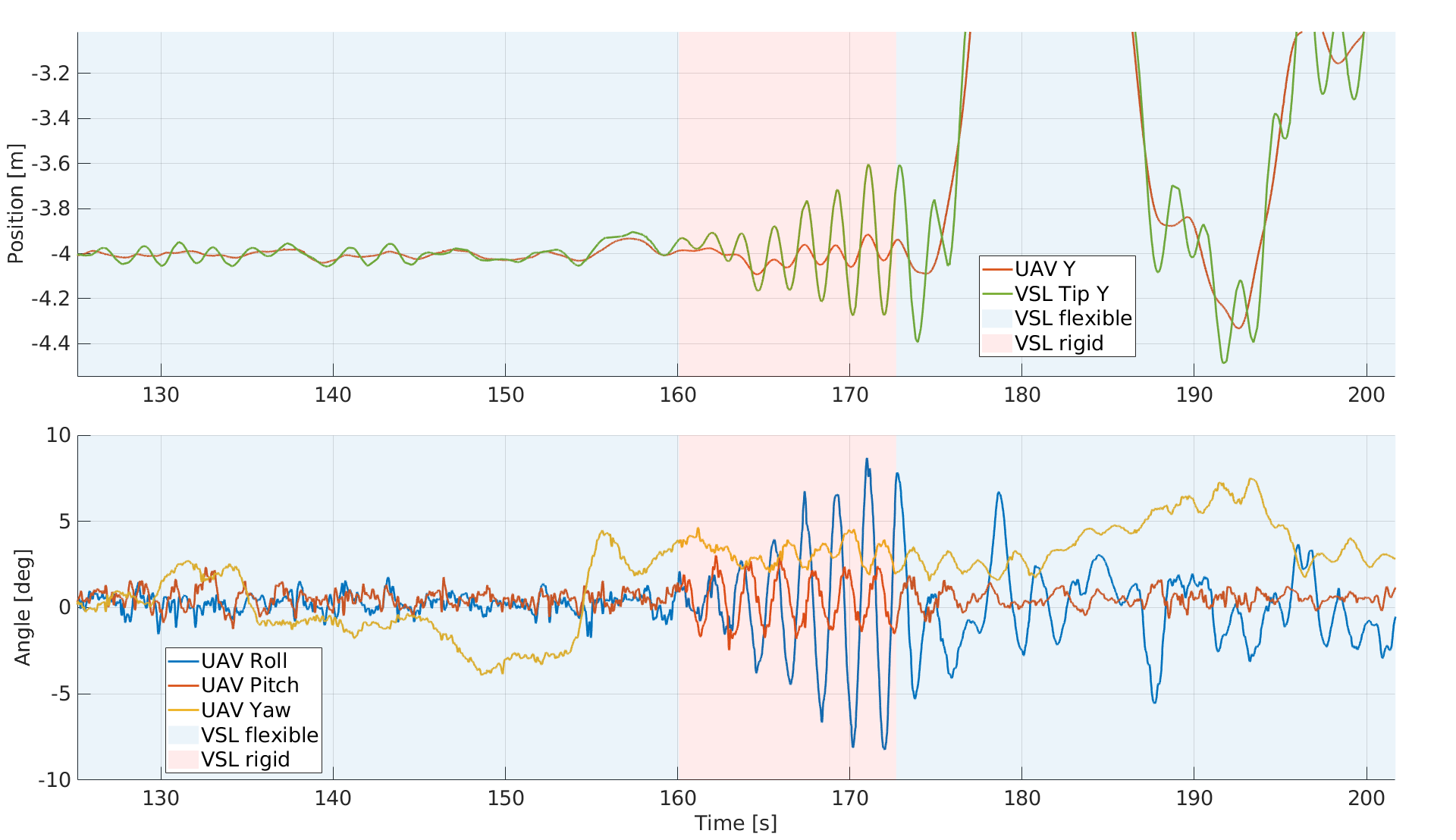}
    \caption{Detailed area from Figure \ref{fig:exp_hits} to focus on the coupled and uncontrolled behave while VSL is rigid. Transition to flexible state absorbed the oscillations.}
    \label{fig:hitting_rigid_detail}
\end{figure}

As shown in Fig. \ref{fig:exp_hits}, external impacts are applied to the VSL tip while the UAV hovered in position mode. During the flexible configuration (blue area in Fig. \ref{fig:hitting_flex_detail}), the tip exhibits oscillations that are not transmitted to the UAV, indicating effective isolation of disturbance energy. In contrast, when the VSL is rigid (red area in in Fig. \ref{fig:hitting_rigid_detail}), coupling increases and small position deviations appear on the UAV. Attitude deviations are larger and even increasing over time when VSL is rigid as shown in Fig. \ref{fig:hitting_flex_detail}.

The second plot highlights this coupling effect during the rigid configurations. The UAV attitude shows larger and increasing deviations over the time, due to the direct transmission from the link. Instead, the transitions to flexible attenuates the oscillations, confirming that the compliant configuration acts as a passive damper, dissipating disturbance energy and preserving UAV stability.

\begin{figure}[!ht]
    \centering
    \includegraphics[width=0.6\linewidth]{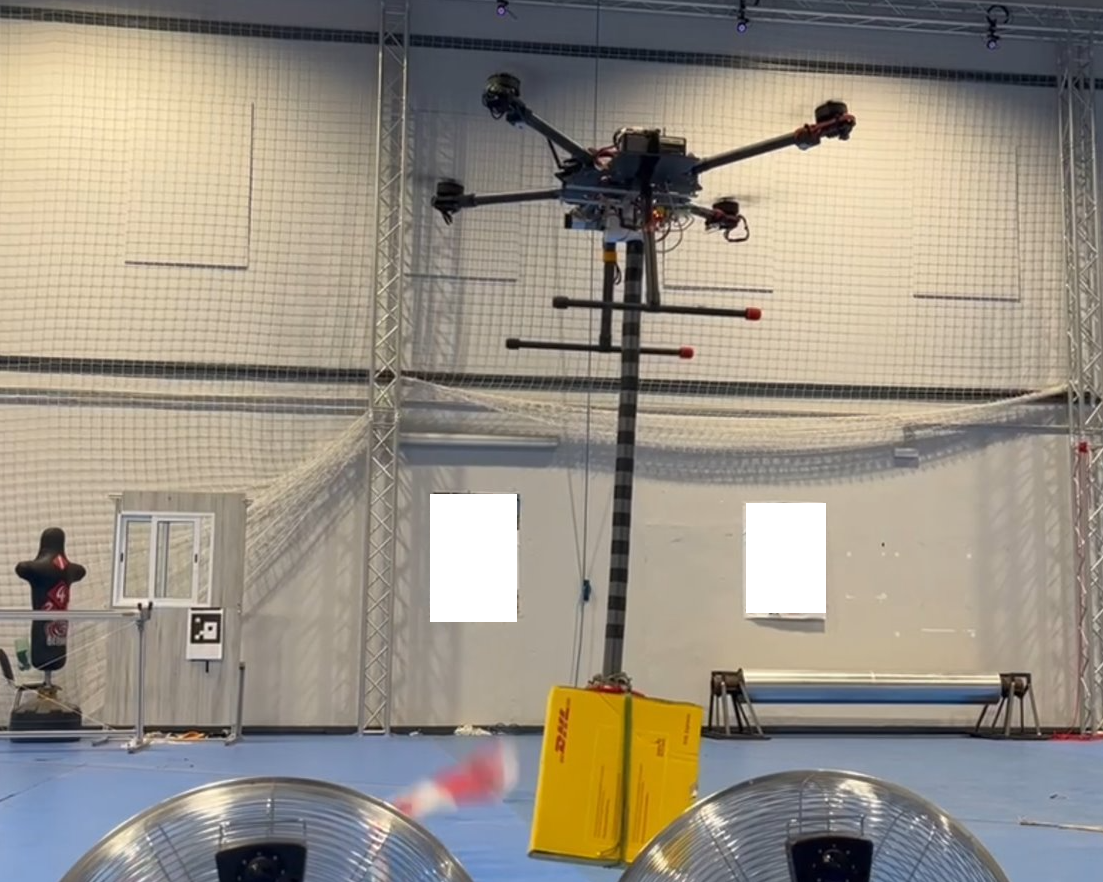}
    \caption{Response to continuous aerodynamic disturbance at the VSL tip. Flexible reduces coupling to the UAV. Rigid increases transmitted motion.}
    \label{fig:fan_setup}
\end{figure}

Next experiment analyzes the behavior of the VSL against aerodynamic forces generated by two fans pointing to the VSL tip while the UAV hovers in position mode\footnote{Wind as external disturbance: \href{https://www.youtube.com/watch?v=7V1Ghp5BtLc}{youtube.com/watch?v=7V1Ghp5BtLc}}. Figure \ref{fig:fan_setup} shows the setup in use from the video recorded performing this evaluation. An empty cardboard box is attached to the tip to make the perturbation more noticeable.

\begin{figure}[!ht]
    \centering
    \includegraphics[width=1.0\linewidth]{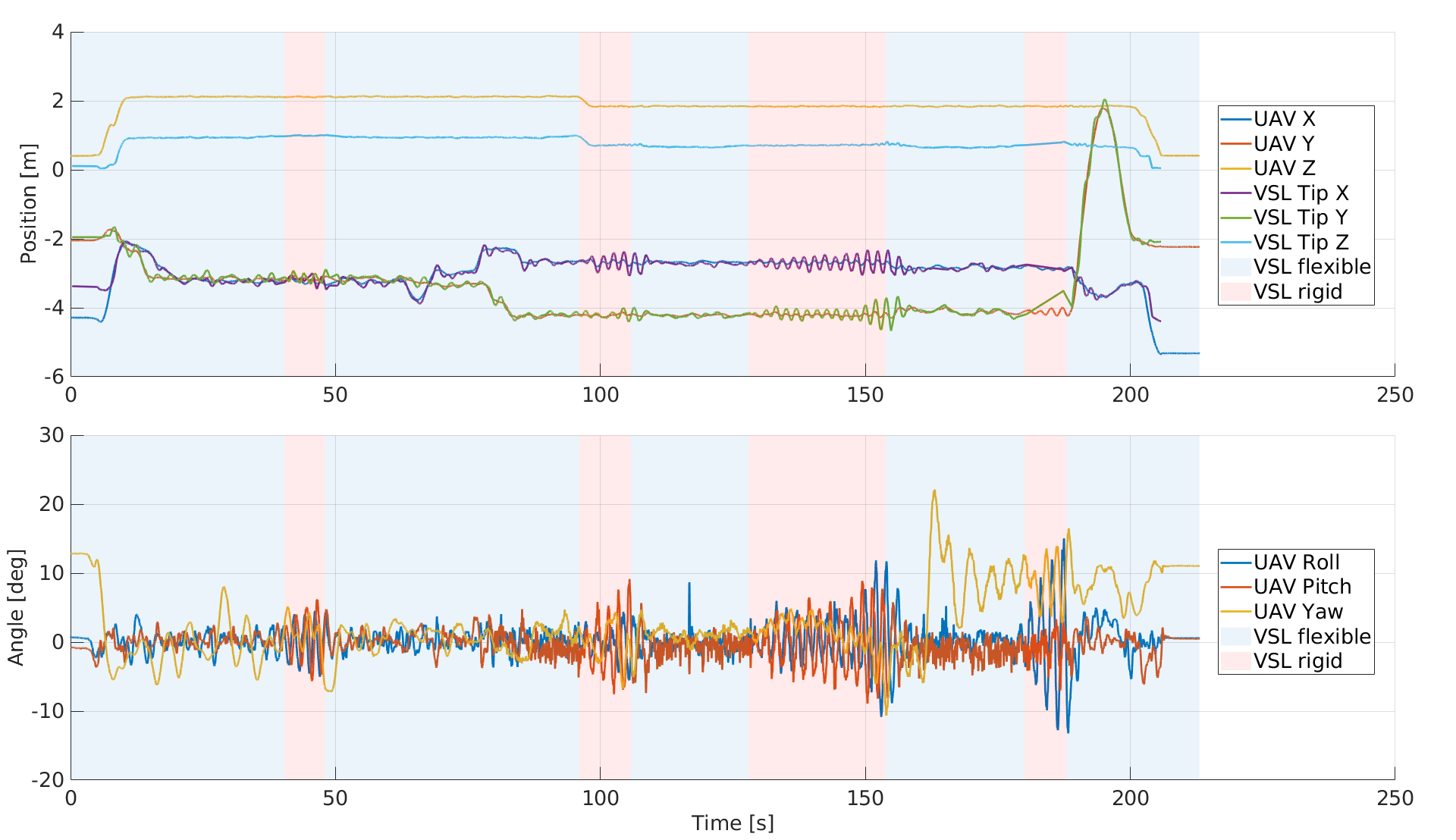}
    \caption{UAV and VSL tip response to external perturbation generated by fans aiming to VSL tip under different stiffness configurations. The first plot represents UAV and VSL tip positions. The second plot shows the the UAV attitude. The VSL stiffness states are highlighted  with blue and red background.}
    \label{fig:exp_fan}
\end{figure}

As shown in the results in Figure \ref{fig:exp_fan}, the VSL tip oscillates while UAV remains stable with the VSL in flexible configuration, dissipating the continuous low force disturbance. Using the rigid configuration, the VSL tip shows a larger angular and position deviation that is transmitted mechanically to the UAV position and attitude.

\begin{figure}[!ht]
    \centering
    \includegraphics[width=1.0\linewidth]{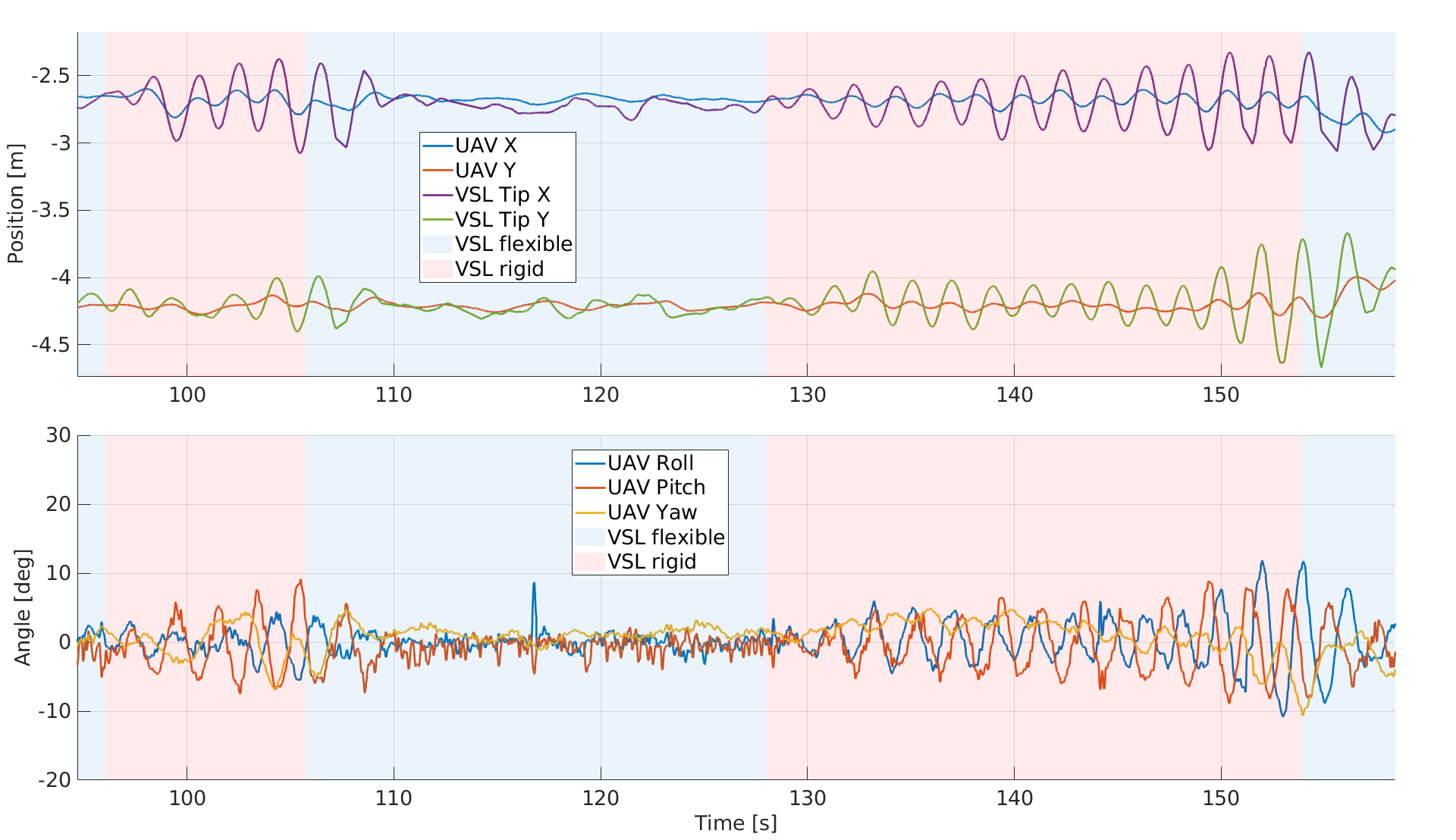}
    \caption{Detailed area from Figure \ref{fig:exp_fan} to display the noticeable oscillations of the VSL during the use of rigid configuration.}
    \label{fig:wind_rigid_detail}
\end{figure}

Figure \ref{fig:wind_rigid_detail} shows a detailed area from Figure \ref{fig:exp_fan} where the oscillations during the rigid state are noticeable in position and attitude. The blue area shows how the flexible state allows to dissipate that energy, keeping the UAV pose stable.

Results from both experiments confirm that the VSL, changing the stiffness, can dynamically adapt the coupling between the UAV and the payload, providing adaptability between compliant isolation and rigid precision depending on the task.

\subsection{Aerial manipulation}
\subsubsection{VSl standalone}
The following experiments evaluate the use of the VSL alone, without the manipulator attached, performing a pick-and-place task where the VSL tip acts as the end-effector. As the previous experiments, the UAV is remotely controlled by a pilot flying in position, closing the control loop with the tracking data from the Motion Capture system.

\begin{figure}[!ht]
    \centering 
    \includegraphics[width=1.0\linewidth]{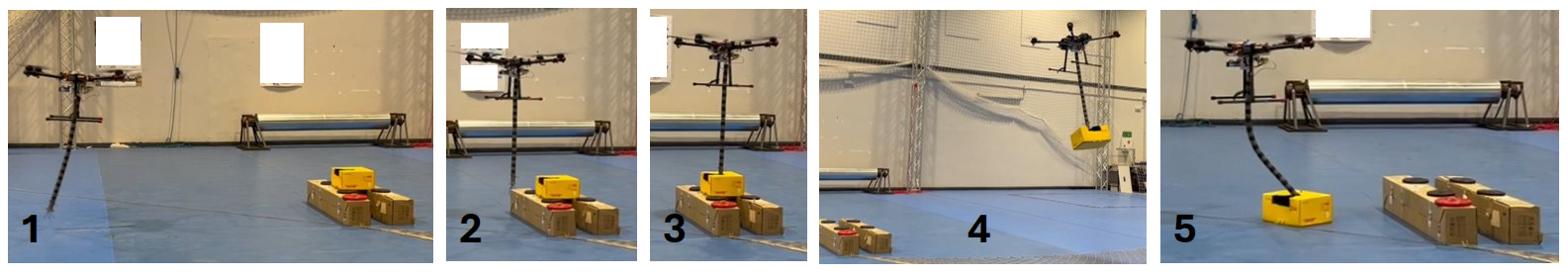}
    \caption{Full sequence combining flex/rigid VSL state for aerial manipulation. Pic 1 shows the UAV appraching to the pickup point (VSL flex). VSL gets rigid and slide the tip in the box on pic 2 and 3. VSL lift the box on pic 4 and gets flex before start the motion. Pic 5 shows the UAV releasing the box on the ground to finish the operation.}
    \label{fig:exp_seq_no_arms}
\end{figure}

As the Figure \ref{fig:exp_seq_no_arms} shows, this setup does not include any payload at the tip of the VSL. The target parcel has hole that allows the VSL tip to slide in and lift the box. The box contains a 2kg weight plate inside.

\begin{figure}[!ht]
    \centering
    \includegraphics[width=1.0\linewidth]{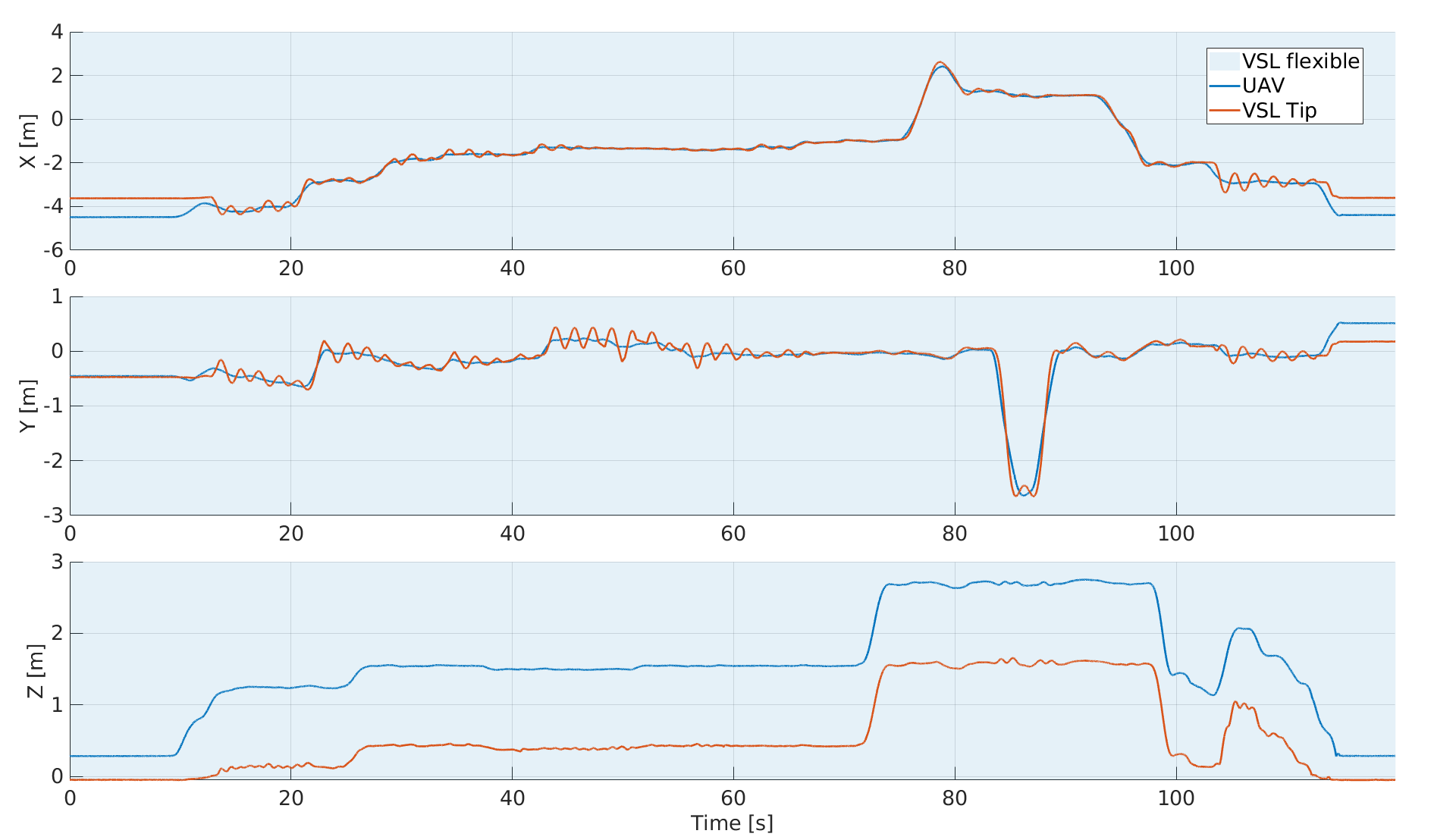}
    \caption{Pickup, transport, and release operation with flexible VSL.}
    \label{fig:exp_no_arms_flex}
\end{figure}

\begin{figure}[!ht]
    \centering
    \includegraphics[width=1.0\linewidth]{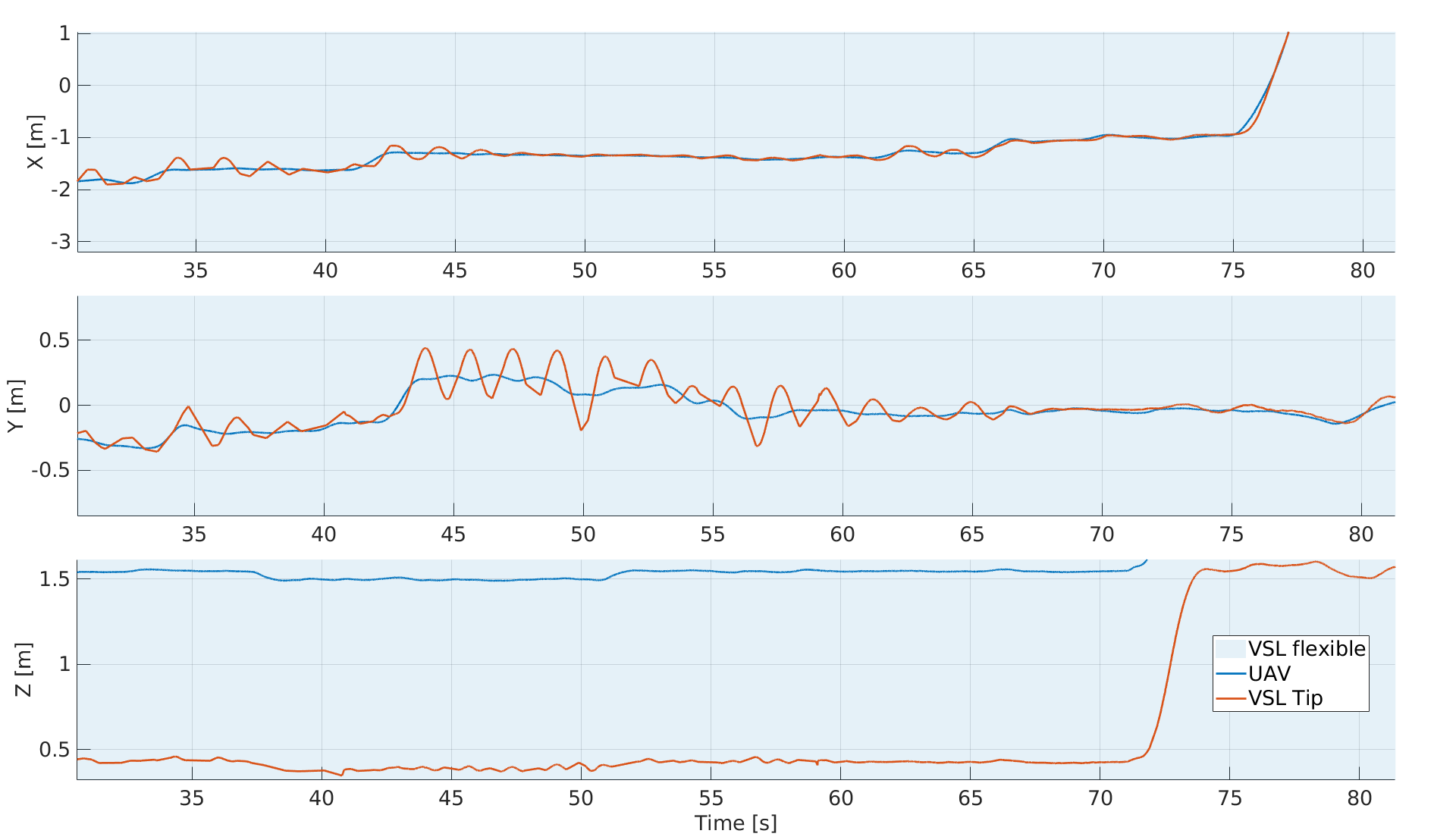}
    \caption{Pick-up box step from Fig. \ref{fig:exp_no_arms_flex}. Relevant details highlighting the oscillations on the VSL tip.}
    \label{fig:no_arms_flex_detail}
\end{figure}

On the first trial, the operation is performed with the VSl in flexible configuration\footnote{Aerial Manipulation with the tip of the VSL, flexible configuration: \href{https://www.youtube.com/watch?v=phHVo9y0DbA}{youtube.com/watch?v=phHVo9y0DbA}}. Figure \ref{fig:exp_no_arms_flex} shows the whole operation, plotting UAV pose and VSL tip position. The VSL tip oscillation is noticeable, specially during the pickup step (\ref{fig:no_arms_flex_detail}), when the pilot attempt to align and insert the link into the box. This step can be recognized since it is done before the UAV increase its height after sliding the VSL tip inside the box.

The next experiment will repeat the operation but combining rigid and flexible configurations\footnote{Aerial Manipulation with the tip of the VSL, combining flexible/rigid configurations: \href{https://www.youtube.com/watch?v=-hg23OeMazM}{youtube.com/watch?v=-hg23OeMazM}}. The rigid configuration is used to reduce oscillations on pickup/release step, i.e. when the pilot requires more accuracy. The flexible configuration is useful to minimize coupling of the payload with the UAV during the transportation to place the UAV in pickup/release target points.

\begin{figure}[!ht]
    \centering
    \includegraphics[width=1.0\linewidth]{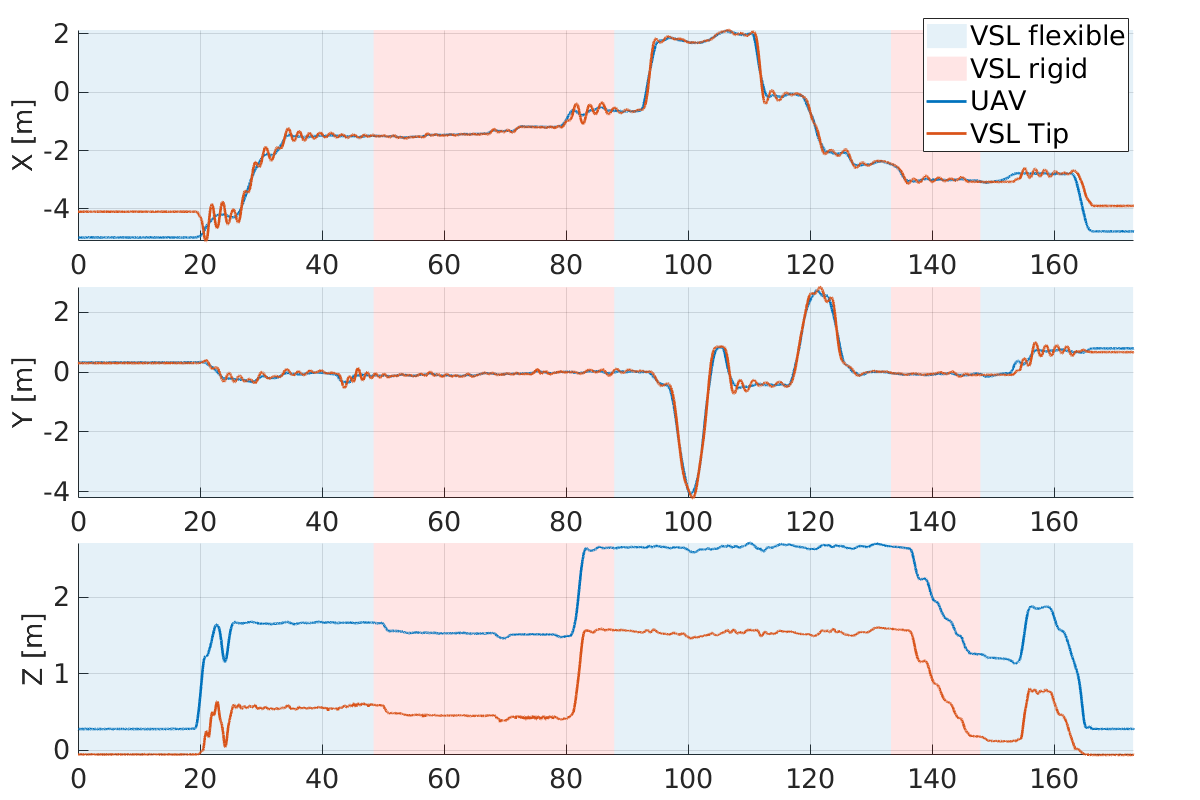}
    \caption{Pickup, transport, and release operation with combined VSL configurations. Load-cell signal shown for reference}
    \label{fig:exp_no_arms_combo}
\end{figure}

\begin{figure}[!ht]
    \centering
    \includegraphics[width=1.0\linewidth]{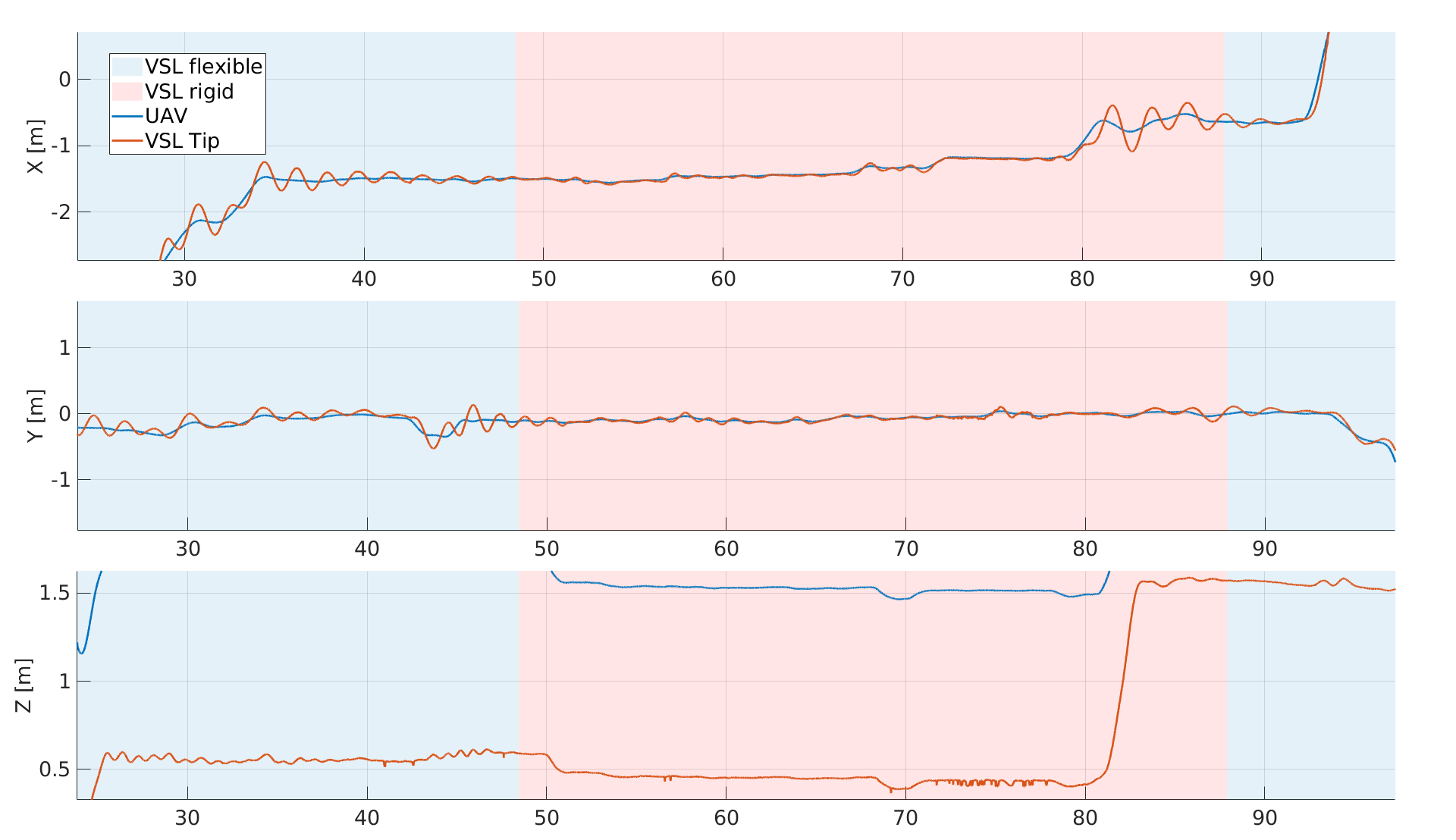}
    \caption{Pick-up box step from Fig. \ref{fig:exp_no_arms_combo}. Relevant details highlighting the oscillations on the VSL tip.}
    \label{fig:exp_no_arms_combo_detail}
\end{figure}

The area shown in Fig. \ref{fig:exp_no_arms_combo_detail} confirms a significant reduction in VSL tip oscillations while the rigid VSL is in use, allowing a more stable pick up operation. 

Although the full task could be completed with a flexible link, the oscillations during the pickup step increase difficulty. The combination of stiffness states thus offers a useful balance between compliance and control accuracy. A real field scenario could include wind and higher positioning error that could make the operation much harder.

\subsubsection{LiCAS integration}

\begin{figure}[!ht]
    \centering
    \includegraphics[width=1.0\linewidth]{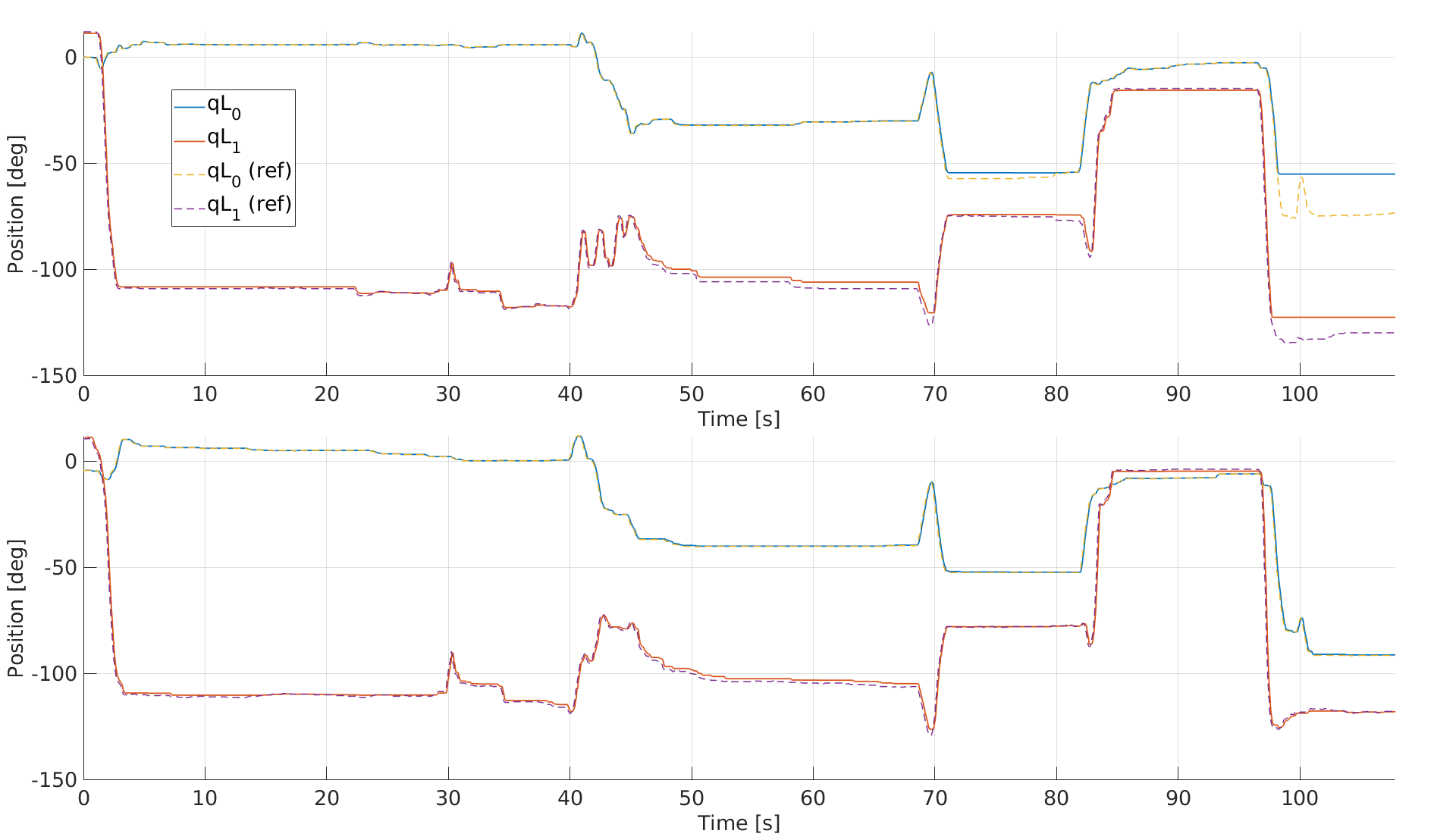}
    \caption{LiCAS teleoperated following references commanded from Ground Station with leader manipulator}
    \label{fig:exp_arms_teleop}
\end{figure}

The final experiments evaluate the integration of the teleoperated LiCAS dual-arm manipulator mounted at the VSL tip for parcel transportation tasks. This manipulator weights 2Kg. It is remotely teleoperated during the experiments by a human operator though a leader/follower architecture as it was depicted in Figure \ref{fig:teleoperation_device}. The operator will move the leader manipulator as a puppet and joints positions are send to the follower as joint references\footnote{Teleoperation test for box grabbing/release: \href{https://www.youtube.com/watch?v=FkNgtGKNmHI}{ youtube.com/watch?v=FkNgtGKNmHI}} (Fig. \ref{fig:exp_no_arms_combo}). Figure \ref{fig:exp_arms_teleop} shows the use of this control scheme. 

\begin{figure}[!ht]
    \centering
    \includegraphics[width=0.6\linewidth]{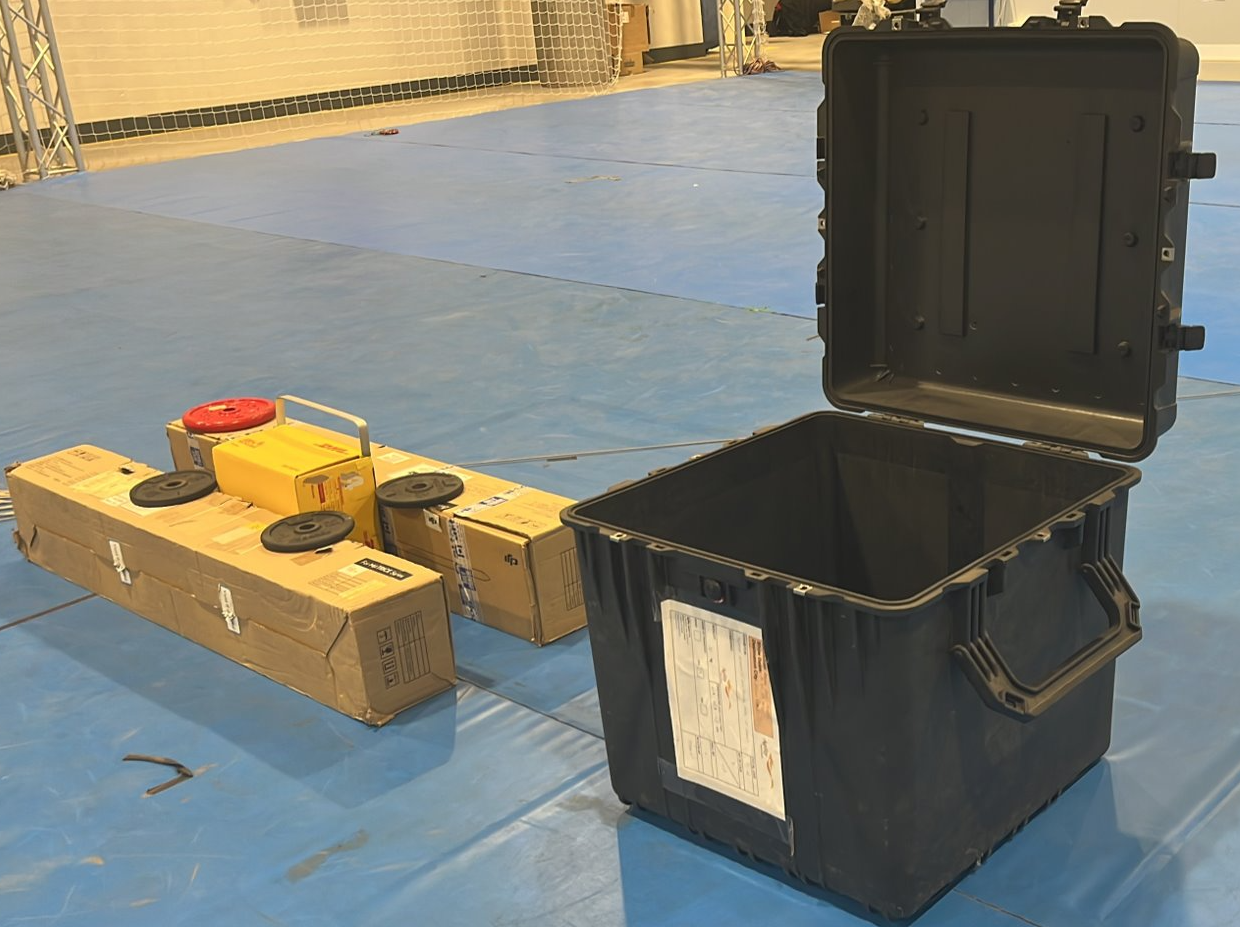}
    \caption{Yellow box (left) to be transported and Pelican case (right) to be used as the destination where the box will be delivered.}
    \label{fig:boxes_arms}
\end{figure}

\begin{figure}[!ht]
    \centering
    \includegraphics[width=1.0\linewidth]{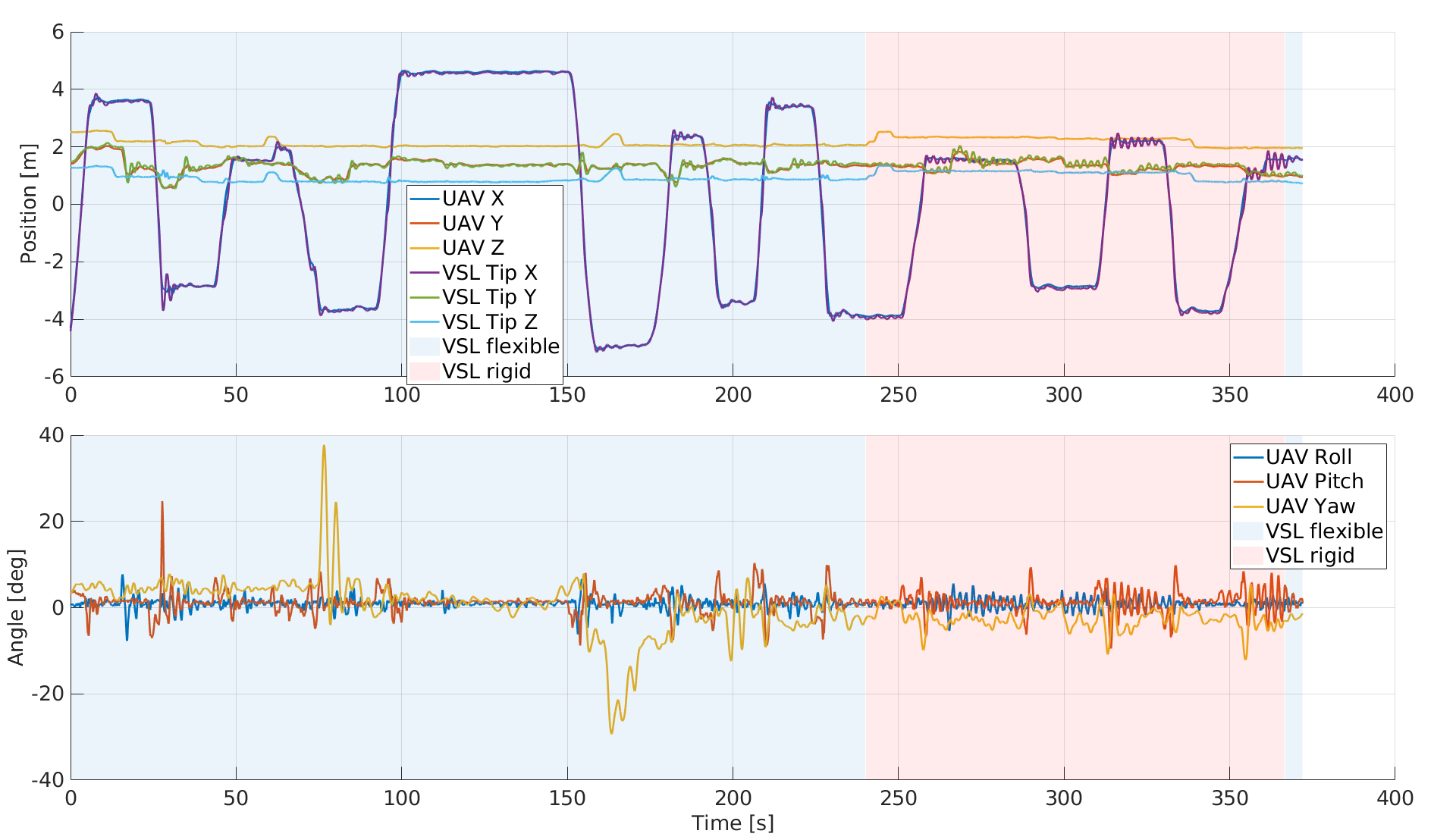}
    \caption{Position data from aerial manipulator and VSL tip while testing the effects of manipulator colliding in motion with a box}
    \label{fig:exp_arms_hit}
\end{figure}

Before performing pick-and-place operations, the aerial manipulator was subjected to light collisions with a stationary Pelican case 0370 (Fig. \ref{fig:boxes_arms}) to evaluate the response to contact\footnote{LiCAS collision test: \href{https://www.youtube.com/watch?v=Wv8rzdZj2Js}{youtube.com/watch?v=Wv8rzdZj2Js}} (Fig. \ref{fig:exp_no_arms_combo}). As Fig. \ref{fig:exp_arms_hit} shows, with the VSL in the flexible state, oscillations induced by impacts quickly dissipated, whereas in the rigid state the disturbances propagated to the UAV, causing larger attitude deviations. The rigidity was slightly lower in this test, unlike in the rest, for safety reasons.

\begin{figure}[!ht]
    \centering
    \includegraphics[width=1.0\linewidth]{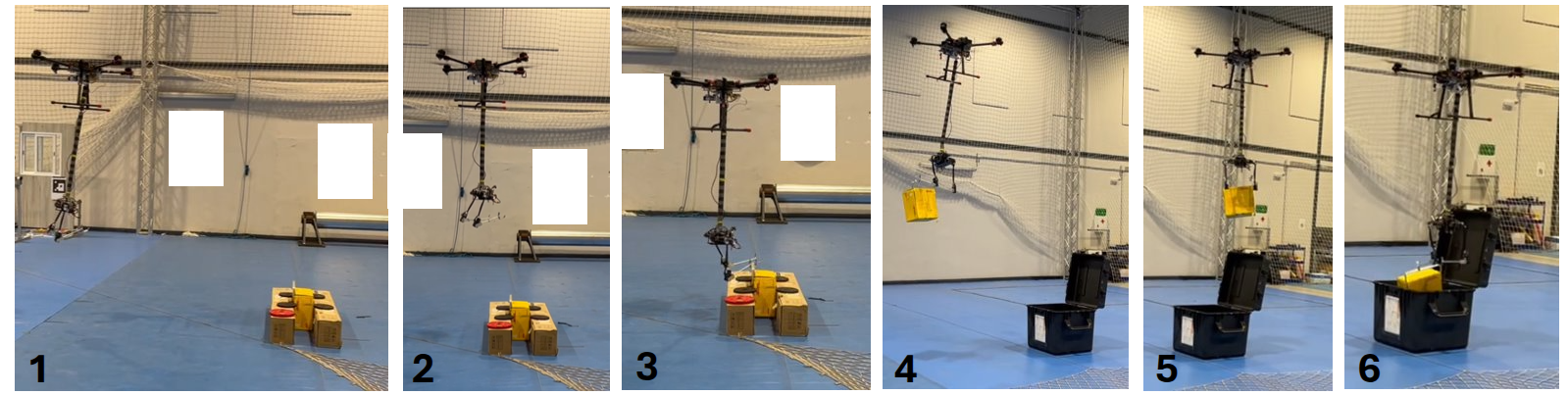}
    \caption{Full sequence combining flexible/rigid VSL state for aerial manipulation with LiCAS attached at the end of the VSL. The UAV goes to pickup point with VSL flex on pic 1. Pics 2 and 3 shows the pickup operation with a rigid VSL. Pic 4 shows the motion to the releasing point (VSL flex). The aerial manipulator delivers the box in the Pelican case on pics 5 and 6 using a rigid VSL.}
    \label{fig:exp_seq_arms}
\end{figure}

For the last experiment, the teleoperated manipulator is used to transport a 600 g box into a Pelican case (Fig. \ref{fig:boxes_arms}). Following the same pickup–transport–release sequence as before (Fig. \ref{fig:exp_seq_no_arms}), the UAV will approach to pickup point to grab the box, will move to the release target point and will deliver the box inside. Figure \ref{fig:exp_seq_arms} display the full sequence by steps

When operated with the flexible VSL (Fig. 22), oscillations at the VSL tip were observed but remained within stable limits. Combining stiffness modes (Fig. 23) produced limited improvement, likely due to the additional mass of the manipulator reducing the relative effect of stiffness variation.” 

\begin{figure}[!ht]
    \centering
    \includegraphics[width=1.0\linewidth]{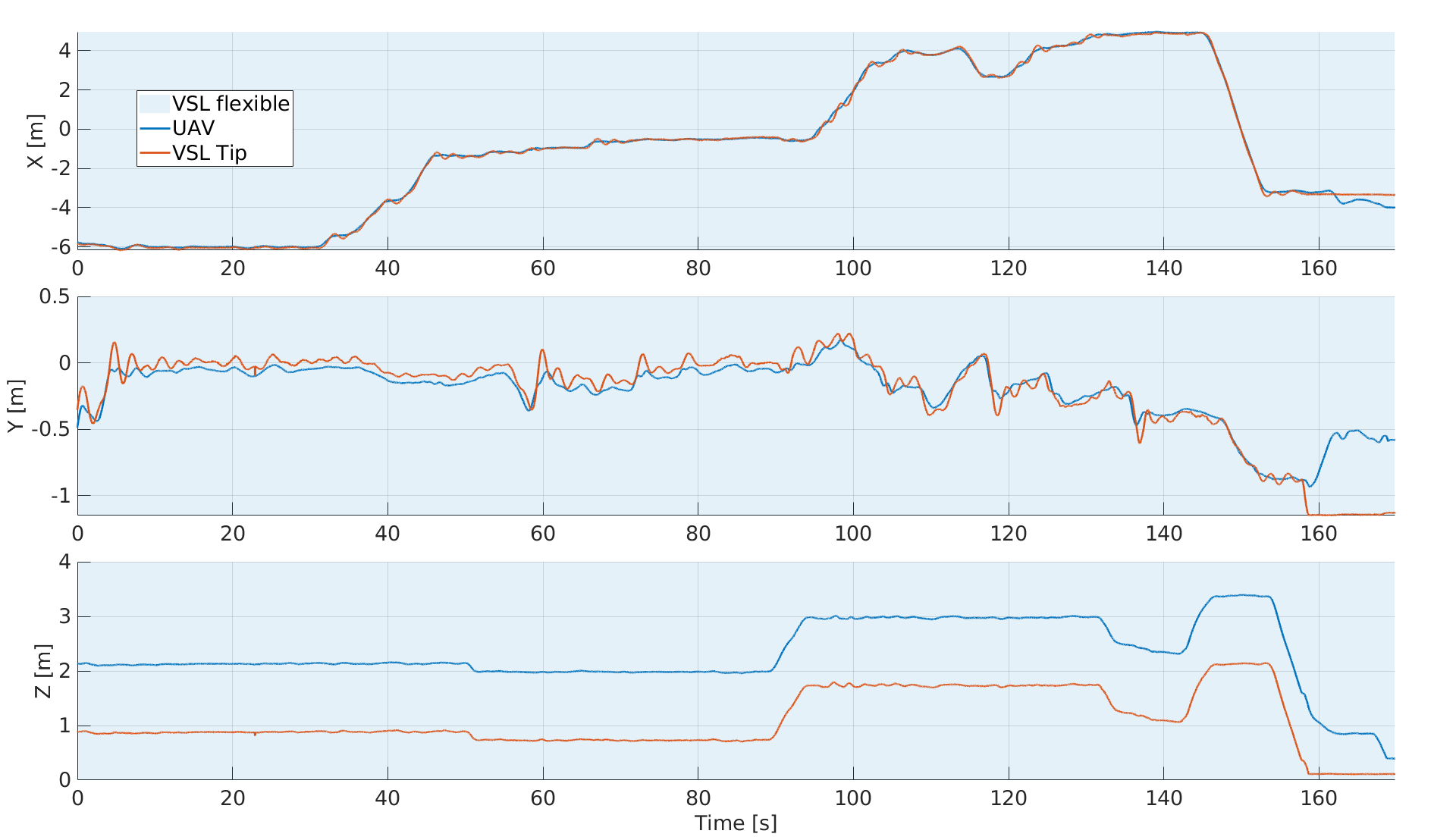}
    \caption{Position data of UAV and manipulator (VSL tip) during parcel transportation with flexible VSL.}
    \label{fig:exp_arms_flex}
\end{figure}

The operation is first evaluated using the VSL with the flexible configuration\footnote{Manipulation with LiCAS dual-arm, flexible configuration: \href{https://www.youtube.com/watch?v=gwNTxSH98fE}{youtube.com/watch?v=gwNTxSH98fE}} (Fig. \ref{fig:exp_no_arms_combo}) (Figure \ref{fig:exp_arms_flex}). As it is expected, oscillations appear at the VSL tip. The weight of the manipulator appears to produce smaller oscillations than those shown in the experiment with the VSL standalone (Fig. \ref{fig:exp_no_arms_flex}).

\begin{figure}[!ht]
    \centering
    \includegraphics[width=1.0\linewidth]{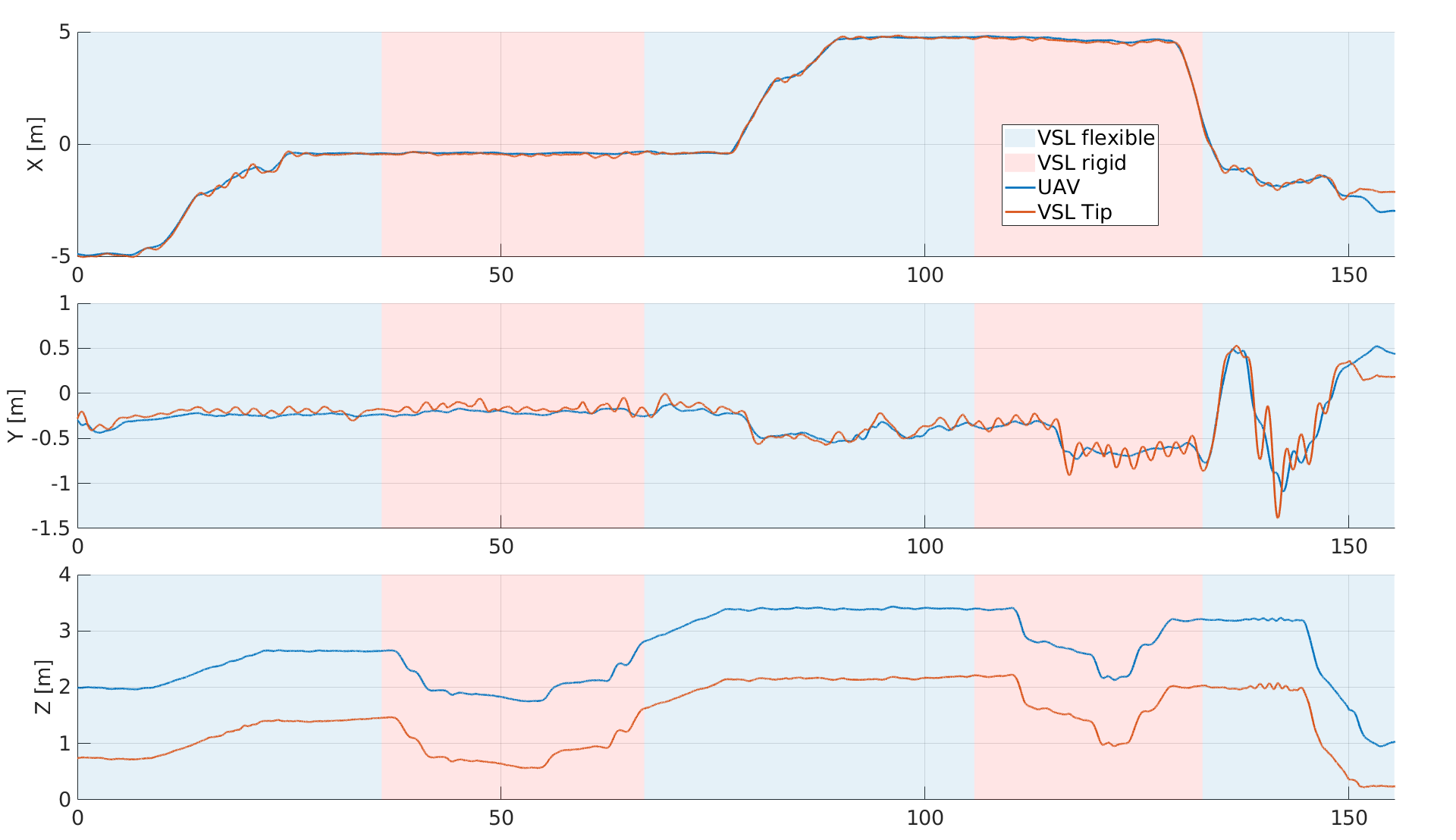}
    \caption{Position data combining flexible/rigid VSL configuration for parcel transportation. Load-cell signal shown for reference.}
    \label{fig:exp_arms_combo}
\end{figure}

In this case, the combination of VSL configurations\footnote{Manipulation with LiCAS dual-arm, combination of flexible/rigid configurations: \href{https://www.youtube.com/watch?v=fuD47SdPyMk}{youtube.com/watch?v=fuD47SdPyMk}} (Fig. \ref{fig:exp_no_arms_combo}) (Fig. \ref{fig:exp_arms_combo}) produces limited improvement, likely due to the additional mass of the manipulator reducing the relative effect of stiffness variation.

Overall, these results confirm that VSL enhances the versatility of aerial manipulation systems. While flexibility improves disturbance rejection and operational safety, adjustable stiffness becomes particularly beneficial when manipulating lightweight payloads or performing precise tasks at the end-effector. 

%% file: sections/z_conclusion.tex
\section{CONCLUSIONS AND FUTURE WORK}\label{sec:conclusions}
This work presented the integration of a Variable Stiffness Link (VSL) into a long-reach aerial manipulation system. The proposed link allows the mechanical coupling between the aerial vehicle and the payload to be adjusted in real time, enabling the system to switch between compliant and rigid configurations depending on the task requirements.

The experimental evaluation validated the expected behavior of the VSL in both configurations. Operating in the flexible state, the VSL effectively isolated external disturbances and attenuated oscillations transmitted to the aerial platform. In the rigid state, it improved accuracy during manipulation task. Transition time between configurations was approximately 7.8 s and remained consistent even with a 2 kg payload, confirming the robustness of the mechanism.

The successful integration of the VSL with the LiCAS dual-arm manipulator demonstrated the feasibility of using adjustable stiffness for long-reach aerial teleoperation. The results indicate that such adaptability can enhance operational safety and precision in tasks involving physical interaction or payload handling.

Future work will focus on increasing stiffness range and miniaturization of the actuation module, integrating stiffness feedback into autonomous control loops and expanding to multi-rope configurations and cooperative aerial manipulation. Additional studies will evaluate user experience and performance improvements in teleoperated and semi-autonomous modes.

%% file: sections/zz_acknowledgement.tex
\section{ACKNOWLEDGMENT}\label{sec:ack}
The authors would like to thank José Manuel Morón Romero and Rocío Gómez Zafra from the GRVC Robotics Lab for their collaboration and support during the experiments.